\pdfoutput=1

\documentclass[11pt]{article}

\usepackage{authblk}
\usepackage[]{acl}

\usepackage{times}
\usepackage{latexsym}

\usepackage[T1]{fontenc}

\usepackage[utf8]{inputenc}

\usepackage{microtype}
\usepackage{booktabs}
\usepackage{multirow}
\usepackage{graphicx}
\usepackage{algorithm}
\usepackage{xspace}
\usepackage{algpseudocode}
\usepackage{array}
\usepackage{amsmath}
\usepackage{float}
\usepackage{multirow}
\usepackage{tikz}
\usepackage{verbatim}
\usetikzlibrary{arrows}
\usepackage{makecell}
\usepackage{subfigure}
\usepackage[inline]{enumitem}
\usepackage[normalem]{ulem}
\usepackage[rgb,dvipsnames]{xcolor}
\usepackage{mathtools}
\usepackage{nameref}
\usepackage{pgfplots}
\usepackage{hyperref}
\usepackage{authblk}

\makeatletter
\newcommand\Autoref[1]{\@first@ref#1,@}
\def\@throw@dot#1.#2@{#1}
\def\@set@refname#1{
    \edef\@tmp{\getrefbykeydefault{#1}{anchor}{}}%
    \xdef\@tmp{\expandafter\@throw@dot\@tmp.@}%
    \ltx@IfUndefined{\@tmp autorefnameplural}%
         {\def\@refname{\@nameuse{\@tmp autorefname}s}}%
         {\def\@refname{\@nameuse{\@tmp autorefnameplural}}}%
}
\def\@first@ref#1,#2{%
  \ifx#2@\autoref{#1}\let\@nextref\@gobble
  \else%
    \@set@refname{#1}
    \@refname~\ref{#1}
    \let\@nextref\@next@ref
  \fi%
  \@nextref#2%
}
\def\@next@ref#1,#2{%
   \ifx#2@ and~\ref{#1}\let\@nextref\@gobble
   \else, \ref{#1}
   \fi%
   \@nextref#2%
}
\makeatother

\title{Generating Gender Alternatives in Machine Translation}

\author[$\S$]{\textbf{Sarthak Garg}}
\author[$\P$]{\textbf{Mozhdeh Gheini}\thanks{\; Work done during an internship at Apple.}\hspace{1.2mm}}
\author[$\S$]{\textbf{Clara Emmanuel}}
\author[$\S$]{\textbf{Tatiana Likhomanenko}}
\author[$\S$]{\authorcr\textbf{Qin Gao}{\thanks{\; Equal senior contribution.}}\hspace{1.2mm}}
\author[$\S$]{\textbf{Matthias Paulik\textbf{$^{\dag}$}}}
\affil[$\S$]{Apple Inc.}
\affil[ ]{\texttt {\{sarthak\_garg,clara\_e,antares,qingao,mpaulik\}@apple.com}}
\affil[$\P$]{Information Sciences Institute, University of Southern California}
\affil[ ]{\texttt {gheini@isi.edu}}
\begin{document}

\maketitle{}

\begin{abstract}
Machine translation (MT) systems often translate terms with ambiguous gender (e.g., English term ``the nurse'') into the gendered form that is most prevalent in the systems' training data (e.g., ``enfermera'', the Spanish term for a female nurse). This often reflects and perpetuates harmful stereotypes present in society. 
With MT user interfaces in mind that allow for resolving gender ambiguity in a frictionless manner, we study the problem of generating \textit{all} grammatically correct gendered translation alternatives. 
We open source train and test datasets for five language pairs and establish benchmarks for this task.
Our key technical contribution is a novel semi-supervised solution for generating alternatives that integrates seamlessly with standard MT models and maintains high performance without requiring additional components or increasing inference overhead.
\end{abstract}
\section{Introduction and Related Work}  
Gender\footnote{``gender'' in this work refers to binary grammatical gender, and not social gender (male, female, nonbinary). Please refer to \S\nameref{sec:limits} for a detailed discussion.} biases present in train data are known to bleed into natural language processing (NLP) systems, resulting in dissemination and potential amplification of those biases \cite{sun-etal-2019-mitigating}. Such biases are often also the root cause of errors. A machine translation (MT) system might, for example, translate \textit{doctor} to the Spanish term \textit{médico} (masculine) instead of \textit{médica} (feminine), given the input ``The doctor asked the nurse to help her in the procedure'' \cite{stanovsky-etal-2019-evaluating}. To avoid prescribing wrong gender assignment, MT systems need to disambiguate gender through context. When the correct gender cannot be determined through context, providing multiple translation alternatives that cover all valid gender choices is a reasonable approach.

Numerous prior works have focused on producing correctly gendered translations given contextual gender ``hints'', such as ``to help her'' in the example above \cite{stanovsky-etal-2019-evaluating,saunders-byrne-2020-reducing,stafanovics-etal-2020-mitigating,Costa-jussà_Escolano_Basta_Ferrando_Batlle_Kharitonova_2022,saunders-etal-2022-first,renduchintala-etal-2021-gender,bentivogli-etal-2020-gender,currey-etal-2022-mt}. In contrast, the problem of generating all valid and grammatically correct gendered translations has seen far less attention \cite{Kuczmarski2018GenderAwareNL, johnson2020, DBLP:journals/corr/abs-2309-03175}.

Consider the example: ``The secretary was angry with the boss.'' 
 The gender of both \textit{secretary} and \textit{boss} remain ambiguous in the absence of additional context: both entities can take either gender. However, and to the best of our knowledge, all existing approaches \cite{Kuczmarski2018GenderAwareNL, johnson2020, DBLP:journals/corr/abs-2309-03175,10.1145/3600211.3604675} for producing different gendered translations operate on ``sentence-level'', instead of on ``entity-level'': they only allow two sentence-level alternatives to surface, in which  both \textit{secretary} and \textit{boss} are either masculine or feminine:
\begin{itemize}[noitemsep,topsep=5pt,leftmargin=.15in]
    \small
    \item {\color{brown}secretary}, {\color{brown} boss}: {\color{brown}El secretario} estaba {\color{brown}enojado} con {\color{brown}el jefe}.\footnote{Gendered translations in Spanish. Brown and teal represent masculine and feminine genders respectively.}
    \item {\color{teal}secretary}, {\color{teal} boss}: {\color{teal}La secretaria} estaba {\color{teal}enojada} con {\color{teal}la jefa}.
\end{itemize}

In this work, we introduce a novel approach that operates on entity-level, i.e., it generates four alternatives corresponding to all grammatically valid combinations of gender choices for both entities:

\begin{itemize}[noitemsep,topsep=5pt,leftmargin=.15in]
    \small
    \item {\color{brown}secretary}, {\color{brown} boss}: {\color{brown}El secretario} estaba {\color{brown}enojado} con {\color{brown}el jefe}.
    \item {\color{brown}secretary}, {\color{teal} boss}: {\color{brown}El secretario} estaba {\color{brown}enojado} con {\color{teal}la jefa}.
    \item {\color{teal}secretary}, {\color{brown} boss}: {\color{teal}La secretaria} estaba {\color{teal}enojada} con {\color{brown}el jefe}.
    \item {\color{teal}secretary}, {\color{teal} boss}: {\color{teal}La secretaria} estaba {\color{teal}enojada} con {\color{teal}la jefa}.
\end{itemize}

When integrated with a proper user interface, our approach provides users with the freedom to choose gender for each entity. We posit that any such system should meet the following practical quality criteria, making the problem challenging:
\begin{itemize}[leftmargin=.15in]
\item{Alternatives should not be produced when the gender can be inferred from the sentence context, e.g., ``She is a boss'' should only produce the feminine translation ``Ella es \textcolor{teal}{una jefa}''.}
\item {All alternatives should maintain grammatical gender agreement. Phrases like ``\textcolor{brown}{El} \textcolor{teal}{secretaria}'' or  ``\textcolor{teal}{secretaria} estaba \textcolor{brown}{enojado}'' should not be produced as they break gender agreement by using different gendered forms for the same entity.}
\item{Alternatives should differ \textit{only} in gender inflections and not general wording, formality, etc., as any such differences can potentially encode bias.}
\end{itemize}

This paper presents several key contributions towards studying the task of generating entity-level alternatives, meeting the above quality criteria:
\begin{itemize}[leftmargin=.15in]
    \item Producing entity-level alternatives for $n$ gender-ambiguous entities requires generating $2^{n}$ different translations. We propose an efficient approach that reduces the problem to generating a \emph{single} structured translation where ``gender-sensitive phrases'' are grouped together and aligned to corresponding ambiguous entities. 
    \item 
    We open source train datasets \footnote{\url{to_be_released}}
    for this task for 5 language pairs and establish supervised baselines. We extend an existing test set for this task: GATE \cite{10.1145/3600211.3604675} from 3 to 6 language pairs and open source the extended set.
    \item We develop a semi-supervised approach that leverages pre-trained MT models or large language models (LLMs) for data augmentation. 
    Models trained on augmented data outperform the supervised baselines and can also generalize to language pairs not covered in the train sets.
\end{itemize}

\section{Entity-Level Gender Alternatives} \label{sec:concepts-rewrite}

Our key insight for efficiently generating entity-level gender alternatives is to reduce the problem to generating a single translation with embedded \emph{gender structures} and their \emph{gender alignments}.

Consider our previous example: ``The secretary was angry with the boss.'' We want to generate the following entity-level alternatives:
\begin{itemize}[noitemsep,topsep=5pt,leftmargin=.15in]
    \small
    \item {\color{brown}secretary}, {\color{brown} boss}: {\color{brown}El secretario} estaba {\color{brown}enojado} con {\color{brown}el jefe}.
    \item {\color{brown}secretary}, {\color{teal} boss}: {\color{brown}El secretario} estaba {\color{brown}enojado} con {\color{teal}la jefa}.
    \item {\color{teal}secretary}, {\color{brown} boss}: {\color{teal}La secretaria} estaba {\color{teal}enojada} con {\color{brown}el jefe}.
    \item {\color{teal}secretary}, {\color{teal} boss}: {\color{teal}La secretaria} estaba {\color{teal}enojada} con {\color{teal}la jefa}.
\end{itemize}

Since we constraint the alternatives to only differ in gender inflections, we can instead produce a single translation with gender-sensitive phrases grouped together as gender structures, shown in $\big(\big)$:
\begin{gather*}
\big(\begin{smallmatrix} \text{\color{brown}El secretario}\\ \text{\color{teal}La secretaria} \end{smallmatrix}\big) \text{ } \text{\small estaba } \big(\begin{smallmatrix} \text{\color{brown}enojado}\\ \text{\color{teal}enojada} \end{smallmatrix}\big) \text{ } \text{\small con } \big(\begin{smallmatrix} \text{\color{brown}el jefe}\\ \text{\color{teal}la jefa} \end{smallmatrix}\big)
\end{gather*}

All alternatives can be derived from this single translation by choosing either the masculine or feminine form in each gender structure. However, doing this naively can give us invalid alternatives that break gender agreement, for example:
\begin{itemize}
\small
\centering
\item[] {\color{brown} El secretario} estaba {\color{teal} enojada}  con {\color{brown} el jefe}
\end{itemize}
$\big(\begin{smallmatrix} \text{\color{brown}El secretario}\\ \text{\color{teal}La secretaria} \end{smallmatrix}\big)$ and $\big(\begin{smallmatrix} \text{\color{brown}enojado}\\ \text{\color{teal}enojada} \end{smallmatrix}\big)$ correspond to the same entity \emph{secretary} and cannot have different gender choices. By having gender alignments between each gender structure in the translation and its corresponding gender-ambiguous entity in the source, we can deduce which gender structures are linked together and need to be consistent with each other.

Let $x = x_1 \ldots x_n$ be the source sentence containing $n$ tokens and let $G_a \subseteq \{1 \ldots n\}$ represent the set of indices of gender-ambiguous entities in~$x$. We aim to produce a translation $y_S$:
\begin{gather}
y_S = y_1 \ldots \big(\begin{smallmatrix} {\color{brown}M_1}\\ {\color{teal}F_1} \end{smallmatrix}\big) \ldots \big(\begin{smallmatrix} {\color{brown}M_k}\\ {\color{teal}F_k} \end{smallmatrix}\big) \dots y_m,
\end{gather}
containing a set of gender structures $S = \{S_1 \ldots S_k\}$ where $S_i\coloneqq \big(\begin{smallmatrix} {\color{brown}M_i}\\ {\color{teal}F_i} \end{smallmatrix}\big)$ is the $i^{th}$ gender structure. Translation $y_S$ is a sequence of two types of elements: $\{y_1 \ldots y_m\} = y_S \setminus S$ are regular tokens that do not change based on the gender of any entity in $G_a$ and $\color{brown}M_*$/$\color{teal}F_*$ are the masculine and feminine inflected forms of the phrases that do change based on the gender of an entity in $G_a$. Gender alignments can then be formally defined as a one-to-many mapping from $G_a$ to $S$. An ambiguous entity is aligned to a gender structure $\big(\begin{smallmatrix} \text{\color{brown}M}\\ \text{\color{teal}F} \end{smallmatrix}\big)$ iff the correct inflection form ({\color{brown}M} or {\color{teal}F}) in the translation depends on the gender of the entity. In our example, \emph{secretary} is aligned to $\big(\begin{smallmatrix} \text{\color{brown}El secretario}\\ \text{\color{teal}La secretaria} \end{smallmatrix}\big)$, $\big(\begin{smallmatrix} \text{\color{brown}enojado}\\ \text{\color{teal}enojada} \end{smallmatrix}\big)$, and \emph{boss} is aligned to $\big(\begin{smallmatrix} \text{\color{brown}el jefe}\\ \text{\color{teal}la jefa} \end{smallmatrix}\big)$.
Given the translation with gender structures $y_S$ and gender alignments, alternatives corresponding to any combination of gender assignments of ambiguous entities can be easily derived as follows: for all ambiguous entities with male gender assignment, choose the male form for their aligned gender structures. Similarly, for all entities with female assignments, choose the female form for their aligned gender structures.

\begin{table*}
\small
\resizebox{\textwidth}{!}{
\begin{tabular}{>{\raggedright\arraybackslash}m{5cm}|>{\raggedright\arraybackslash}m{6cm}|>{\raggedright\arraybackslash}m{4cm}}
\textbf{Source annotations} & \textbf{Target annotations} & \textbf{Alignment annotations}\\
\hline
The \textbf{lawyer} fought to keep his \textbf{child}, who is a gangster, safe from the \textbf{judge}.

\textbf{lawyer} $\rightarrow$ Masculine

\textbf{child} $\rightarrow$ Gender-Ambiguous

\textbf{judge} $\rightarrow$ Gender-Ambiguous
&
El abogado luchó para mantener a su $\big(\begin{matrix} \text{\color{brown}hijo}\\ \text{\color{teal}hija} \end{matrix}\big)$, que es $\big(\begin{matrix} \text{\color{brown}un}\\ \text{\color{teal}una} \end{matrix}\big)$ gángster, a salvo $\big(\begin{matrix} \text{\color{brown}del juez}\\ \text{\color{teal}de la jueza} \end{matrix}\big)$. &
\textbf{child} $\rightarrow \big(\begin{matrix} \text{\color{brown}hijo}\\ \text{\color{teal}hija} \end{matrix}\big), \big(\begin{matrix} \text{\color{brown}un}\\ \text{\color{teal}una} \end{matrix}\big)$

\textbf{judge} $\rightarrow \big(\begin{matrix} \text{\color{brown}del juez}\\ \text{\color{teal}de la jueza} \end{matrix}\big)$
\\
\hline

\end{tabular}
}
\vspace{-0.5em}
\caption{English–Spanish annotation example. \textit{lawyer}, \textit{child} and \textit{judge} are the annotated entities. \textit{child} and \textit{gangster} refer to the same entity and \textit{child} is selected as the head-word. \textit{lawyer} is marked as masculine because of the co-referring pronoun \textit{his} and is translated to the masculine form: \textit{El abogado}. \textit{child} and \textit{judge} are gender-ambiguous leading to gender structures in the translation (middle column) and gender alignments (rightmost column).}
\vspace{-1.5em}
\label{tab:annotation-steps}
\end{table*}
\section{Datasets}
To build and evaluate systems producing alternatives, we prepare train and test sets containing gender structures and gender alignment annotations.
\subsection{Test data}
\label{subsec:test_data}
We evaluate our models on a combination of two existing test sets that test complementary aspects:
\begin{itemize}[noitemsep,topsep=5pt,leftmargin=.15in]
\item GATE \cite{10.1145/3600211.3604675} has source sentences with at least $1$ and at most $3$ gender-ambiguous entities with their entity-level alternatives satisfying our quality criteria. It evaluates the system on cases where alternatives \emph{should} be produced.
\item MT-GenEval \cite{currey-etal-2022-mt} contains sentences with entities whose gender can be inferred from the sentence context and are not ambiguous. This set is helpful for evaluating cases where alternatives \emph{should not} be produced. 
\end{itemize}

These two test sets have different annotation formats and guidelines. In order to unify them, we ask annotators to review and post-edit existing annotations using the following guidelines:
\begin{enumerate}[noitemsep,topsep=5pt,leftmargin=.15in]
\item \textbf{Marking gendered words}: First, all words in the source referring to entities (people/animals) that can have masculine or feminine grammatical genders are marked.
\item \textbf{Gender ambiguity annotation}: Next, if multiple words refer to the same entity, a head word is selected among them. We guided the annotators to pick the one that acts the most like the subject as the head word. For each head word, if its gender can be inferred from the grammatical context, such as co-referring male/female pronouns, it is marked as such. If no gender can be inferred, the gender is marked as ambiguous. We only rely on grammatical sentence context and not on external knowledge/common gender associations of names/proper nouns. \autoref{app:masculine_generics} discusses how our annotation guidelines handle the problem of masculine generics \citep{piergentili-etal-2023-gender}, where masculine nouns/pronouns can be used to refer to ambiguous or collective entities.
\item \textbf{Gender aware translation}: Finally, we ask the annotators to translate the source sentence. Entities without any ambiguity must be translated into the correct gender. If the translation depends on the gender of the ambiguous entities in the source, gender structures and gender alignments are annotated.
\end{enumerate}

\autoref{tab:annotation-steps} explains the process with the help of an example annotation. We prepare this unified test set for $6$ language pairs:
English to German, French, Spanish, Portuguese, Russian, and Italian.\footnote{We extend the original GATE corpus, which only includes English to Spanish, French, and Italian.}
\subsection{Train data}
\label{subsec:train_data}
We open source train data containing samples in the same format as the test set to ensure reproducibility and to encourage development of supervised/semi-supervised systems for producing alternatives. In contrast to the test sets, which are created via human annotation, we rely on an automatic data augmentation approach (see \autoref{app:train_data} for details) to create train data at scale. The source sentences for the train sets are sampled from Europarl \cite{koehn-2005-europarl}, WikiTitles \cite{TIEDEMANN12.463}, and WikiMatrix \cite{schwenk-etal-2021-wikimatrix} corpora. The train data are partitioned into two different sets:
\begin{itemize}[noitemsep,topsep=5pt,leftmargin=.15in]
\item \textbf{G-Tag} contains source sentences with head words for all entities with their gender ambiguity label: Masc., Fem. or Ambiguous.
\item \textbf{G-Trans} contains gender-ambiguous entities in the source sentences, gender structures in the translations and gender alignments. 
\end{itemize}

To the best of our knowledge, this is the first large-scale corpus that contains gender ambiguities and how they effect gendered forms in the translation. We release these sets for $5$ language pairs: English to German, French, Spanish, Portuguese, and Russian. G-Tag contains $\sim12$k sentences and G-Trans contains $\sim50k$ sentence pairs per language pair.
Detailed statistics of the train and test sets can be found in \autoref{app:dataset-details}.
\section{Training MT Models to Generate Gender Structures and Alignments} \label{subsec:student}
We first present how to train MT models that produce gender structures and alignments, assuming parallel data enriched with gender structures and alignments (for example, G-Trans) is available. We then describe a novel data augmentation pipeline that can enrich any regular parallel corpora with gender structures and alignments.

Given a source sentence $x = \{x_1 \ldots x_n\}$, translation $y_S$ containing gender structures, and gender alignments $A$, we want to train the MT model to generate $y_S, A | x$. Let's assume that $y_S$ contains $k$ gender structures and $A = \{a_1 \ldots a_k\}$ where $a_i$ represents the source token aligned to the $i^{th}$ gender structure. We serialize each gender structure in $y_S$ into a sequence of tokens as follows:
\begin{equation*}
\big(\begin{smallmatrix} \text{\color{brown}M}\\ \text{\color{teal}F} \end{smallmatrix}\big) \rightarrow \texttt{BEG }\text{\small \color{brown}M}\texttt{ MID }\text{\small\color{teal}F}\texttt{ END}
\end{equation*}
where \texttt{BEG}, \texttt{MID}, and \texttt{END} are special tokens. The model is then trained to produce gender structures in the form of this sequence. 

\citet{garg-etal-2019-jointly} introduced a technique to train MT models to jointly generate translations and word-alignments. We use their approach to learn generation of gender alignments. Let $m_1 \ldots m_k$ denote the positions of the \texttt{MID} tokens of the gender structures. A specific cross-attention head is chosen and supervised to learn gender alignments. Let $n$ and $m$ denote the lengths of the source and the serialized target respectively and let $P_{m\times n}$ denote the attention probability distribution computed by the selected head. We train the model with regular cross entropy and an additional \emph{alignment loss}:
\begin{equation*}
L = L_{\text{cross-ent}} -\frac{\lambda}{k}\sum_{i=1}^{k}log(P_{m_i, a_i})    
\end{equation*}
where $\lambda$ is a scaling factor. This added loss term encourages the attention head to place more probability mass on the aligned source token when generating the \texttt{MID} token belonging to that token's gender structure.
During inference, the gender alignment for the $i^{th}$ gender structure can be computed as:
\begin{equation*}
a_i = \operatorname*{argmax}_{s\in\{x_1 \ldots x_n\}} P_{m_i,s}
\end{equation*}

This model can generate gender structures and alignments without any additional inference overhead. Then, using the procedure described in \autoref{sec:concepts-rewrite}, all entity-level alternatives can be easily derived from the model outputs.
{\renewcommand{\arraystretch}{1.3}
\begin{table*}
\small
\resizebox{\textwidth}{!}{
\begin{tabular}{>{\raggedright\arraybackslash}m{3cm}|>{\raggedright\arraybackslash}m{6cm}|>{\raggedright\arraybackslash}m{8cm}}
 & \textbf{Source} & \textbf{Target} \\
\hline
G-Trans dataset &
The \textcolor{blue}{doctor} was angry with the \textcolor{red}{patient}

\textcolor{blue}{doctor} $\rightarrow$ Gender-Ambiguous

\textcolor{red}{patient} $\rightarrow$ Gender-Ambiguous
&
$\big(\begin{matrix} \text{\color{blue}El doctor}\\ \text{\color{blue}La doctora} \end{matrix}\big)$ estaba $\big(\begin{matrix} \text{\color{blue}enojado}\\ \text{\color{blue}enojada} \end{matrix}\big)$ con $\big(\begin{matrix} \text{\color{red}el}\\ \text{\color{red}la} \end{matrix}\big)$ paciente\\
\hline

Fine-tuning bi-text &
\makecell{The doctor\texttt{<M>} was angry with the patient\texttt{<M>} \\ 
The doctor\texttt{<F>} was angry with the patient\texttt{<F>} \\
The doctor\texttt{<M>} was angry with the patient\texttt{<F>} \\
The doctor\texttt{<F>} was angry with the patient\texttt{<M>}} 
&
\makecell{\emph{El doctor} estaba \emph{enojado} con \emph{el} paciente \\
\emph{La doctora} estaba \emph{enojada} con \emph{la} paciente \\
\emph{El doctor} estaba \emph{enojado} con \emph{la} paciente \\
\emph{La doctora} estaba \emph{enojada} con \emph{el} paciente
} \\

\hline
\end{tabular}
}
\vspace{-0.5em}
\caption{Extracting bi-text for fine-tuning from the G-Trans dataset. Each gender-ambiguous token is suffixed with a gender assignment tag: \texttt{<M>/<F>}. With the help of alignments (shown via color coding), the correct gender inflection is selected in the translation. $n$ ambiguous entities can result in $2^n$ different assignments, but we only keep "all-masculine", "all-feminine", and a maximum of $3$ other randomly sampled assignments.}
\vspace{-1em}
\label{tab:gender-assignment-aware-bitext}
\end{table*}}
\section{Data Augmentation Pipeline}
\label{sec:data-aug}
G-Trans dataset provides supervised data to train MT models in the above manner. However, this dataset is small ($50k$ examples per language pair) and has a restrictive domain, limiting the quality of the trained models. We propose a \emph{data augmentation} pipeline that can take any regular parallel corpora (containing high quality but potentially biased translations) and augment the translations with gender structures and alignments whenever there are ambiguities in the source.
\algnewcommand{\LineComment}[1]{\State \(\triangleright\) #1}
\renewcommand{\algorithmicrequire}{\textbf{Input:}}
\renewcommand{\algorithmicensure}{\textbf{Output:}}

\begin{algorithm}[h]
\small
\caption{Data Augmentation Overview}\label{alg:teacher}
\begin{algorithmic}
\Require $x = \{x_1 \ldots x_n\}$ (source sentence) and $y_B=\{y_1 \ldots y_m\}$ (reference translation without gender structures, potentially biased) \\
\LineComment{\textbf{Step 1: }Detect set of gender-ambiguous entities $G_a$ in the source sentence: $G_a \subseteq \{1\ldots n\}$}
\State $G_a \gets \text{GenderAmbiguousEntities}(x)$
\If{$G_a = \phi$}
  \State \textbf{Output:} $x, y_B, \phi$ 
\EndIf \\
\LineComment{\textbf{Step 2: }Transform $y_B$ into an all-masculine $y_M$ and all-feminine $y_F$ translations}
\State $y_{M} \gets \text{argmax } p(y | x, y_B, \text{gender}(x_i) = \text{male } \forall i\in G_a)$
\State $y_{F} \gets \text{argmax } p(y | x, y_B, \text{gender}(x_i) = \text{female } \forall i\in G_a)$\\
\LineComment{\textbf{Step 3: }Combine $y_M$ and $y_F$ into a single translation $y_S$ containing gender structures}
\State Let $y_M = y_1\ldots M_1 \ldots y_j \ldots M_k \ldots y_m$ and 
\State Let $y_F = y_1\ldots F_1 \ldots y_j \ldots F_k \ldots y_m$
\State where $y_{*}$ are the common tokens between $y_M$ and $y_F$ and $\{(M_i, F_i)\text{ | } i\in 1\ldots k\}$ be the $k$ differing phrases.
\State $ y_S \gets \text{group}(y_M, y_F) = y_1 \ldots \big(\begin{smallmatrix} M_1\\ F_1 \end{smallmatrix}\big) \ldots \big(\begin{smallmatrix} M_k\\ F_k \end{smallmatrix}\big) \dots y_m$ \\
\LineComment{\textbf{Step 4: }Align each gender structure $S_i\coloneqq \big(\begin{smallmatrix} M_i\\ F_i \end{smallmatrix}\big)$ to its corresponding ambiguous entity in $G_a$}
\State $A \gets \text{ComputeGenderAlignments}(x, y_S)$
\Ensure $x, y_s, A$
\end{algorithmic}
\end{algorithm}
\autoref{alg:teacher} gives an overview of the main components of the pipeline, which we describe in detail in the following subsections. It consists of first detecting gender-ambiguous entities in the source sentence (\S\ref{subsec:amb-tagging}), followed by transforming the reference translation into all-masculine/all-feminine translations (\S\ref{subsec:struct-gen}, \S\ref{subsec:llm}), condensing those into single translation with gender structures, and finally aligning the gender structures (\S\ref{subsec:align}). 
\subsection{Detecting gender-ambiguous entities}
\label{subsec:amb-tagging}

Traditionally, rule-based methods, which rely on dependency parsing and co-reference resolution, are used to detect gender-ambiguous entities in the source sentence \citep{10.1145/3600211.3604675, habash-etal-2019-automatic}. In contrast, we adopt a data-driven approach. G-Tag dataset contains English source sentences annotated with head-words, which refer to entities with their gender label derived from the grammatical sentence context: ambiguous, masculine, feminine.
Following \citet{alhafni-etal-2022-user}, we fine-tune a (\texttt{BERT}-style) pre-trained language model (PLM) using this dataset to tag each source token with one of the four labels: ambiguous, masculine, feminine, or not a headword.

\subsection{Generating all-masculine/feminine translations using fine-tuned MT models}
\label{subsec:struct-gen}
If ambiguous entities are detected in the source sentence, then the next step is to transform the high-quality but potentially biased reference translation $y_B$ to all-masculine $y_M$ and all-feminine $y_F$ translations. $y_M$ and $y_F$ are equivalent to sentence-level alternatives corresponding to masculine and feminine assignments for all ambiguous entities, respectively. We explore two methods for this task: fine-tuning pre-trained MT models (this subsection) and using LLMs (\autoref{subsec:llm}).

We fine-tune a pre-trained MT model $M$ on a bi-text extracted from the G-Trans dataset. The source sentences of this bi-text contain ambiguous entities tagged as masculine or feminine using \texttt{<M>}/\texttt{<F>} tags, and the target translation has correct gender inflections given the gender tags. \autoref{tab:gender-assignment-aware-bitext} explains this extraction process in detail using an example.

The fine-tuned model $M_{\text{fine-tuned}}$ learns to generate translations with gender inflections in agreement with the gender assignments (\texttt{<M>}/\texttt{<F>}) in the source.
We use \citet{saunders-byrne-2020-reducing}'s lattice rescoring approach to generate $y_M$ and $y_F$. Let $x_M$ and $x_F$ denote source sentences in which all ambiguous entities ($G_a$) have been tagged using \texttt{<M>} and \texttt{<F>} tags, respectively. Let $I(y_B)$ represent the search space consisting only of all possible gender inflection variants of $y_B$. $M_{\text{fine-tuned}}$ is used to decode $y_M$ and $y_F$ over the constrained search space $I(y_B)$:
\begin{equation*}
y_M = \operatorname*{argmax}_{y\in I(y_B)}p_{M_{\text{fine-tuned}}}(y|x_M)
\end{equation*}
\begin{equation*}
y_F = \operatorname*{argmax}_{y\in I(y_B)}p_{M_{\text{fine-tuned}}}(y|x_F)
\end{equation*}

This can be done efficiently using constrained beam search. This procedure guarantees that $y_M$, $y_F$, and $y_B$ differ only in terms of gender inflections, and therefore, $y_M$ and $y_F$ possess the same general translation quality as the reference translation $y_B$. 
\begin{figure}[h]
    \centering
    \includegraphics[width=\columnwidth]{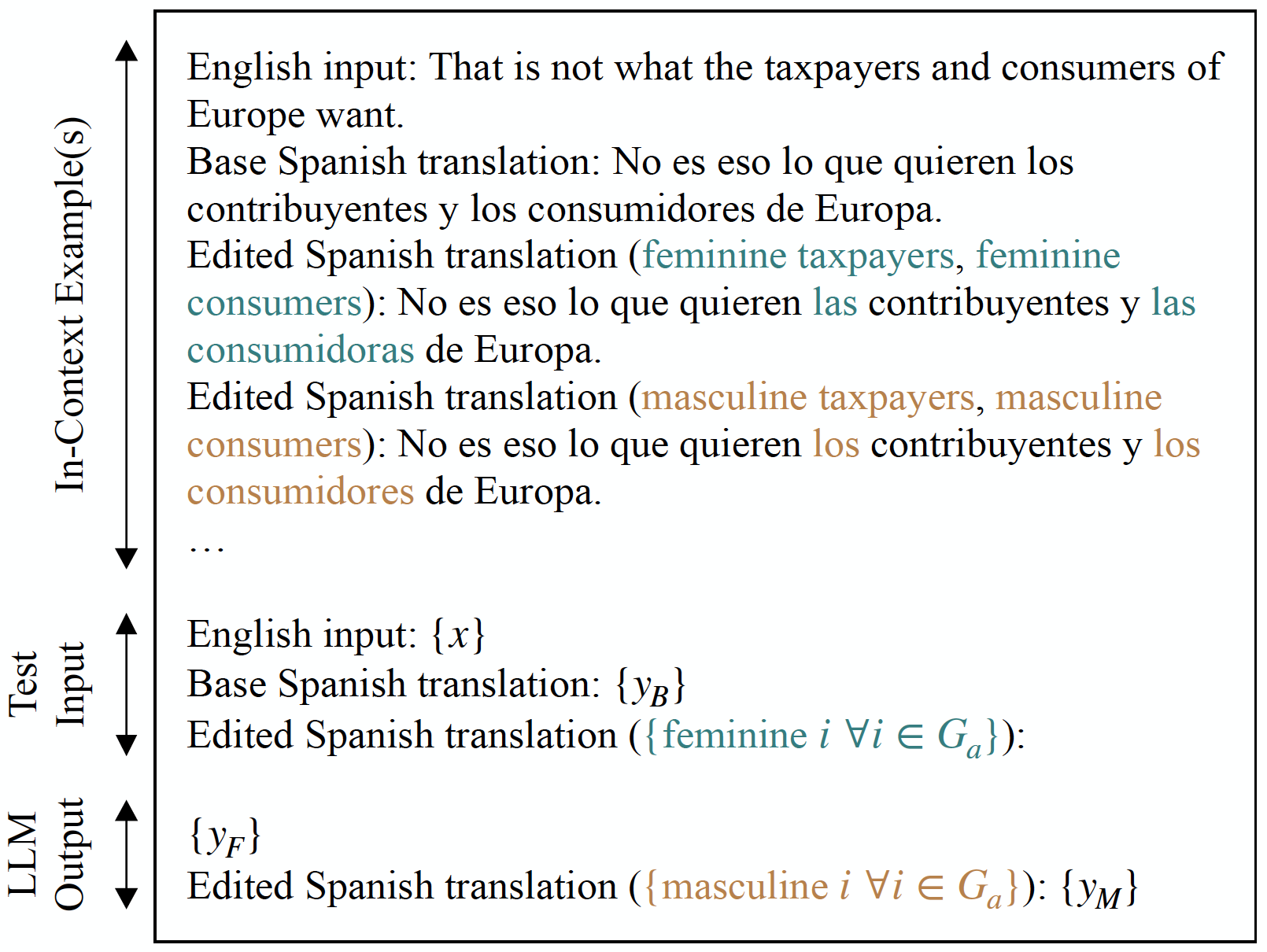}
    \caption{Prompting LLMs using in-context examples to edit the reference translation $y_B$ into all-masculine and all-feminine gender assignments. Multiple in-context examples are used but we illustrate only one here for brevity.}
    \label{fig:icl-editor}
\end{figure}

\subsection{Generating all-masculine/feminine translations using LLMs}
\label{subsec:llm}
LLMs' ability to learn using in-context examples \cite{NEURIPS2020_1457c0d6} provides us with an alternative approach for generating $y_M$ and $y_F$. We can provide selected instances from G-Trans as in-context examples in the prompt to the LLM and have it generate output for a test instance \cite{DBLP:journals/corr/abs-2309-03175}.
Inspired by re-writing literature \cite{vanmassenhove-etal-2021-neutral,DBLP:journals/corr/abs-2102-06788}, we design a prompt that treats the LLM as an editor: it edits/re-writes the given translation $y_B$ to match the provided gender assignments (all-masculine and all-feminine) in the prompt (See \autoref{fig:icl-editor} for an example).

\subsection{Aligning gender structures}
$y_M$ and $y_F$ are combined together in Step 3 as described in \autoref{alg:teacher} to produce a single translation $y_S$ containing gender structures. The final step is to align each gender structure in $y_S$ to an ambiguous entity in the source. We model this as a tagging task and fine-tune a PLM using alignment annotations in the G-Trans dataset.
\label{subsec:align}
\begin{algorithm}[h]
\small
\caption{Alignment Algorithm}\label{alg:alignment}
\begin{algorithmic}
\Require $x = \{x_1 \ldots x_n\}$ (source sentence) and $y_S$ (translation with $k$ gender structures) \\
\State Let $ y_S = y_1 \ldots \big(\begin{smallmatrix} M_1\\ F_1 \end{smallmatrix}\big) \ldots \big(\begin{smallmatrix} M_k\\ F_k \end{smallmatrix}\big) \dots y_m$ \\
\For{$i^{th}$ gender structure $S_i\coloneqq \big(\begin{smallmatrix} M_i\\ F_i \end{smallmatrix}\big)$}
\State Let $|$ be a special marker token
\State $y_A \gets y_1 \ldots M_1 \ldots |M_i| \ldots M_k \ldots y_m$
\State $a_i \gets \text{PLM}(x;y_A)$ \Comment{; denotes concatenation}
\EndFor
\Ensure $A = \{a_i, \forall i \in 1 \ldots k\}$
\end{algorithmic}
\end{algorithm}

Each gender structure is aligned one-by-one as described in \autoref{alg:alignment}. To align the $i^{th}$ gender structure $S_i$, we take $y_M$ and enclose the phrase corresponding to $S_i$ by a special token \texttt{|} to get $y_A$. Then $x$ and $y_A$ are concatenated together and fed to the PLM, which is fine-tuned to tag all the tokens in $x$ as aligned/not-aligned to $S_i$ (See \autoref{fig:alignment} in the Appendix for an example). The gold aligned/not-aligned labels for fine-tuning are extracted from the G-Trans dataset.

{\renewcommand{\arraystretch}{1.1}
\begin{table*}[ht]
\centering
\small
\begin{tabular}{|c|c|cc|c|cc|c|}
\hline
\multirow{2}{1.5cm}{\centering Language Pair} & \multirow{2}{*}{Model} & \multicolumn{2}{c|}{Alternatives Metrics $\uparrow$} & \multirow{2}{*}{$\delta$-BLEU $\downarrow$} & \multicolumn{2}{c|}{Structure Metrics $\uparrow$} & \multirow{2}{2cm}{\centering Alignment $\uparrow$ Accuracy\%}\\
& & Precision\% & Recall\% & & Precision\% & Recall\% & \\ \hline\hline
\multirow{2}{*}{En–De} & Fine-tuned M2M & $\mathbf{94}$ & $89.7$ & $4.7$ & $87.8$ & $91$ & \multirow{2}{*}{$93.7$}\\ 
      & GPT & $91.1$ & $\mathbf{92.7}$ & $\mathbf{2.8}$ & $\mathbf{89.8}$ & $\mathbf{94}$ & \\ \hline
\multirow{2}{*}{En–Es} & Fine-tuned M2M & $\mathbf{95.7}$ & $91.6$ & $3.3$ & $\mathbf{88.1}$ & $\mathbf{93}$ & \multirow{2}{*}{$91.5$}\\ 
      & GPT & $91.5$ & $\mathbf{93.7}$ & $\mathbf{2.7}$ & $84.7$ & $92.7$ & \\ \hline
\multirow{2}{*}{En–Fr} & Fine-tuned M2M & $\mathbf{93.8}$ & $\mathbf{92.5}$ & $3.6$ & $\mathbf{88.1}$ & $92.9$ & \multirow{2}{*}{$92.9$}\\ 
      & GPT & $89.4$ & $91$ & $\mathbf{2.8}$ & $85.8$ & $\mathbf{94.8}$ & \\ \hline
\multirow{2}{*}{En–Pt} & Fine-tuned M2M & $\mathbf{94.8}$ & $\mathbf{94.3}$ & $\mathbf{3.5}$ & $88.3$ & $92.4$ & \multirow{2}{*}{$93.6$}\\ 
      & GPT & $93.8$ & $83.5$ & $5.5$ & $\mathbf{89.6}$ & $\mathbf{95.2}$ & \\ \hline
\multirow{2}{*}{En–Ru} & Fine-tuned M2M & $\mathbf{89.4}$ & $\mathbf{89.3}$ & $\mathbf{5.7}$ & $\mathbf{87}$ & $\mathbf{87.7}$ & \multirow{2}{*}{$93.2$}\\ 
      & GPT & $83.5$ & $58.2$ & $10.6$ & $83.1$ & $85$ & \\ \hline
En–It & Fine-tuned M2M & $95.4$ & $87.9$ & $8.2$ & $79.4$ & $75.3$ & $94.1$\\ \hline      
\end{tabular}
\caption{Data augmentation pipeline results. $\uparrow$ indicates higher-the-better and $\downarrow$ lower-the-better metrics.}
\label{tab:lattice-rescoring-results}
\end{table*}}
\section{Evaluation Metrics} \label{sec:evaluation-metrics}
We evaluate our systems' performance using the following metrics:
\begin{itemize}[noitemsep,topsep=5pt,leftmargin=.15in]
\item \textbf{Alternatives metrics}: These metrics compute the overlap between the set of sentences that have alternatives in the test set and the set of sentences for which the system produces alternatives. This overlap is measured using precision and recall and gives a sense of how often the system produces alternatives and whether it produces them only when needed.
\item \textbf{Structure metrics}: These metrics are computed over the set of sentences for which both the test set and system output contain alternatives. They measure the quality of the generated alternatives by computing the overlap between the gender structures in the reference alternatives and the generated alternatives. The overlap is measured using precision and recall.
\item \textbf{Alignment accuracy}: This is measured as the \% of gender structures that are aligned to the correct source entity and reflects the quality of gender agreement in the generated alternatives.
\item \textbf{$\delta$-BLEU}: Lastly, following \newcite{currey-etal-2022-mt}, to measure the degree of bias towards a gender, we compute \emph{$\delta$-BLEU} as follows: We separate the masculine and feminine forms in gender structures (if any) for the reference and the system output, compute masculine and feminine BLEU scores (using \texttt{sacrebleu} \cite{post-2018-call}), and measure the absolute difference between the two:
\begin{equation*}
\delta\text{-BLEU} = |\text{BLEU}(\hat{y}_m, y_m) - \text{BLEU}(\hat{y}_f, y_f)|
\end{equation*}
Higher $\delta$-BLEU indicates more bias. Mathematical definitions of alternatives and structure metrics can be found in \autoref{app:evaluation metrics}.
\end{itemize}
\section{Experiments and Results}
\label{sec:res}
We will first describe the experimental details and results of our data augmentation pipeline in \ref{subsec:data-aug-details} and \ref{subsec:data-aug-results}. We then present the training details of the MT model generating alternatives end-to-end and how it benefits from data augmentation in \ref{subsec:end2end-model-details} and \ref{subsec:end2end-model-results}. 
\subsection{Data augmentation pipeline details}
\label{subsec:data-aug-details}
The data augmentation pipeline consists of three components: detecting gender-ambiguous entities, generating all-masculine/feminine translations and aligning gender structures.

We build the ambiguous entity detector (\S\ref{subsec:amb-tagging}) by fine-tuning \texttt{xlm-roberta-large} \cite{conneau-etal-2020-unsupervised} using \texttt{transformers} \cite{wolf-etal-2020-transformers}. We use the combined G-Tag dataset across all $5$ language pairs for fine-tuning.

To generate all-masculine/feminine translations, we explore two approaches: fine-tuning pre-trained MT models (\S\ref{subsec:struct-gen}), and using LLMs (\S\ref{subsec:llm}). For the first approach, we fine-tune the \texttt{M2M 1.2B} \cite{10.5555/3546258.3546365} model using \texttt{fairseq} \cite{ott-etal-2019-fairseq}. The model is fine-tuned jointly on bi-text extracted from the G-Trans dataset (as described in \autoref{tab:gender-assignment-aware-bitext}) for all $5$ language pairs. 
The list of gender inflections used for lattice rescoring is collected from Wiktionary \cite{ylonen-2022-wiktextract}
 and inflections present in the G-Trans train and test sets.

For the second approach, we use \texttt{gpt-3.5-turbo} as our LLM and follow the prompt design described in \autoref{subsec:llm} with $6$ in-context examples. We provide additional ablation studies on the number of in-context examples, different prompt designs, and choice of LLM (\texttt{gpt} vs. \texttt{OpenLlama-v2-7B} \cite{openlm2023openllama}) in \autoref{app:llm-ablation}. We find that using more in-context examples helps, but gains are minimal for $>6$. Since LLM decoding does not use lattice rescoring, it is possible that the generated all-masculine/feminine translations differ in more than just gender inflections. To avoid this, we explicitly check the differences and don’t generate gender structures if the differences don’t match any entry in the list of gender inflections.

Lastly, to align gender structures we fine-tune \texttt{xlm-roberta-large} on source, targets, and gender alignments extracted from the G-Trans dataset jointly for all $5$ language pairs. The hyper-parameters for fine-tuning XLM and M2M models are decided based on validation performance on a held-out portion of the train sets and can be found in \Autoref{app:tagger,app:finetuned-m2m,app:aligner}.
\subsection{Data augmentation pipeline results}
\label{subsec:data-aug-results}
The data augmentation pipeline takes source sentences and their reference translations (without gender structures, potentially biased) as inputs. For evaluating the data augmentation pipeline, we feed in the source sentences and their all-masculine reference translations from the test set as inputs. The pipeline returns these translations augmented with gender structures and alignments. We can then compute the evaluation metrics described in \autoref{sec:evaluation-metrics} on the generated gender structures and alignments. \autoref{tab:lattice-rescoring-results} summarizes the results. 

Both M2M and GPT perform mostly on par with the exception of English-Russian, where GPT achieves much lower alternatives recall (58.7 compared to 89.3). The quality of generated gender structures is better for GPT on English-German and English-Portuguese and better for M2M on English-Spanish and English-Russian, as can be seen from the structure metrics. Note that we don’t have any G-Trans data for English-Italian, so the results of the M2M model and the alignment accuracy on English-Italian are purely due to zero-shot generalization of M2M and XLM models \cite{johnson-etal-2017-googles}. Overall, the zero-shot results are comparable to others in terms of alternatives metrics and alignment accuracy but fall behind on structure metrics. The alignment model performs well obtaining $\geq 91\%$ accuracy on all language pairs. 

$\delta$-BLEU depends on both alternatives and structure metrics and can be used as a single metric to compare systems' performance. 
Overall, GPT wins in terms of not relying on any fine-tuning dataset and better performance on English to German, Spanish, and French. Fine-tuning M2M wins in terms of achieving better results on English to Portuguese and Russian and being much more efficient in terms of parameters and inference cost (\texttt{M2M 1.2B} can be fit on a single A100 GPU).

\begin{table*}[!ht]
\centering
\small
\begin{tabular}{|c|c|cc|c|cc|c|c|}
\hline
\multirow{3}{1.5cm}{\centering Language Pair} & \multirow{3}{1.5cm}{\centering Model} & \multicolumn{2}{c|}{\multirow{2}{1.5cm}{\centering{Alternatives Metrics $\uparrow$}}} & \multirow{2}{*}{\multirow{3}{1.1cm}{\centering $\delta$-BLEU $\downarrow$}} & \multicolumn{2}{c|}{\multirow{2}{1.5cm}{\centering Structure Metrics$\uparrow$}} & \multirow{3}{1.5cm}{\centering Alignment Accuracy $\uparrow$ \%} & \multirow{3}{1.5cm}{\centering FLoRes BLEU $\uparrow$}\\
& &  & & & & & & \\
& & P\% & R\% & & P\% & R\% & & \\ \hline\hline
\multirow{3}{*}{En–De} & Vanilla & - & - & $8.6$ & - & - & - & $31.6$ \\
      & Supervised & $74.4$ & $71.5$ & $2.4$ & $\mathbf{55.2}$ & $\mathbf{57.5}$ & $89.1$ & $\mathbf{31.9}$ \\
      & w/ Augmented Data & $\mathbf{86.7}$ & $\mathbf{87.5}$ & $\mathbf{0.8}$ & $48.2$ & $55.6$ & $\mathbf{94.2}$ & $31.6$ \\ \hline
\multirow{3}{*}{En–Es} & Vanilla & - & - & $10.4$ & - & - & - & $\mathbf{26}$\\
      & Supervised & $78.9$ & $77.3$ & $2.8$ & $60.5$ & $60.6$ & $85.2$ & $25.9$ \\
      & w/ Augmented Data & $\mathbf{94.3}$ & $\mathbf{92}$ & $\mathbf{1}$ & $\mathbf{62.4}$ & $\mathbf{66.4}$ & $\mathbf{92.5}$ & $\mathbf{26}$\\ \hline      
\multirow{3}{*}{En–Fr} & Vanilla & - & - & $8.1$ & - & - & - & $\mathbf{46.3}$\\
      & Supervised & $74.5$ & $67.8$ & $3.1$ & $\mathbf{60.7}$ & $61.7$ & $82.1$ & $44.9$\\
      & w/ Augmented Data & $\mathbf{87.3}$ & $\mathbf{86.7}$ & $\mathbf{0.8}$ & $59$ & $\mathbf{67.3}$ & $\mathbf{92.5}$ & $45.8$\\ \hline      
\multirow{3}{*}{En–Pt} & Vanilla & - & - & $12.5$ & - & - & - & $\mathbf{44.6}$\\
      & Supervised & $83.4$ & $82.6$ & $3.1$ & $\mathbf{60}$ & $60.9$ & $86.9$ & $43.7$\\
      & w/ Augmented Data & $\mathbf{92.2}$ & $\mathbf{94.4}$ & $\mathbf{1.1}$ & $59.5$ & $\mathbf{63.5}$ & $\mathbf{94.2}$ & $44.1$\\ \hline      
\multirow{3}{*}{En–Ru} & Vanilla & - & - & $5.3$ & - & - & - & $25.6$\\
      & Supervised & $70.6$ & $54.5$ & $2.4$ & $\mathbf{42}$ & $39.5$ & $83.7$ & $\mathbf{26.4}$ \\
      & w/ Augmented Data & $\mathbf{80.7}$ & $\mathbf{77.2}$ & $\mathbf{1.5}$ & $37.6$ & $\mathbf{39.8}$ & $\mathbf{91}$ & $24.9$\\ \hline      
\multirow{2}{*}{En–It} & Vanilla & - & - & $11.6$ & - & - & - & $\mathbf{27.9}$\\
      & w/ Augmented Data & $93.7$ & $89.4$ & $\mathbf{3.2}$ & $53$ & $50.9$ & $94.6$ & $27.6$\\ \hline      
\end{tabular}
\caption{End-to-end MT model results. P and R denote precision and recall respectively.}
\label{tab:student-results}
\end{table*}

{\renewcommand{\arraystretch}{1.1}
\begin{table}[!h]
\centering
\small
\begin{tabular}{|c|c|c|c|c|c|} \hline
LP & Direction & Model & P\% & R\% & F0.5  \\ \hline\hline
\multirow{4}{*}{En–Es} & \multirow{2}{*}{M$\rightarrow$F}&GATE & $\mathbf{95}$ & $40$ & $0.75$ \\
&  & Ours & $89.6$ & $\mathbf{69.2}$ & $\mathbf{0.85}$ \\ \cline{2-6}
& \multirow{2}{*}{F$\rightarrow$M} & GATE & $\mathbf{97}$ & $50$ & $0.82$ \\
& & Ours & $94.5$ & $\mathbf{73.7}$ & $\mathbf{0.89}$ \\ \hline
\multirow{4}{*}{En–Fr} & \multirow{2}{*}{M$\rightarrow$F} & GATE & $\mathbf{91}$ & $27$ & $0.62$ \\
 & & Ours & $89.3$ & $\mathbf{72.5}$ & $\mathbf{0.85}$ \\ \cline{2-6}
 & \multirow{2}{*}{F$\rightarrow$M} &GATE & $\mathbf{97}$ & $28$ & $0.65$ \\
 & & Ours & $96.1$ & $\mathbf{79.3}$ & $\mathbf{0.92}$ \\ \hline
\multirow{4}{*}{En–It} & \multirow{2}{*}{M$\rightarrow$F} &GATE & $\mathbf{91}$ & $32$ & $0.66$ \\
& & Ours & $78.7$ & $\mathbf{58.8}$ & $\mathbf{0.74}$ \\ \cline{2-6}
& \multirow{2}{*}{F$\rightarrow$M} &GATE & $\mathbf{96}$ & $47$ & $0.79$ \\
 & &Ours & $92$ & $\mathbf{75.1}$ & $\mathbf{0.88}$ \\ \hline
\end{tabular}
\caption{Comparison of data augmentation pipeline using M2M against GATE on M $\rightarrow$ F and F $\rightarrow$ M re-writing. P and R denote precision and recall.}
\label{tab:gate-comparison}
\end{table}}
Finally, \autoref{tab:gate-comparison} compares the performance of our data augmentation pipeline using M2M against GATE's sentence-level gender re-writer on their setup. We use our pipeline to re-write an all-masculine reference into an all-feminine form (M$\rightarrow$F) and vice-versa (F$\rightarrow$M). More details about the setup and evaluation metrics used for this comparison can be found in \autoref{app:gate}. We see significant improvements in recall at the cost of relatively small degradation in precision (except English-Italian). Our system is able to outperform GATE on their proposed F.5 metric on all $3$ language pairs.
\subsection{End-to-end MT model details}
\label{subsec:end2end-model-details}
We train a vanilla multilingual MT model on all $6$ language pairs using parallel corpora from Europarl, WikiMatrix, WikiTitles, Multi-UN \cite{chen-eisele-2012-multiun}, NewsCommentary \cite{barrault-etal-2019-findings} and Tilde MODEL \cite{rozis-skadins-2017-tilde}. We refer to this as \emph{vanilla bi-text}. We evaluate the models on gender-related metrics using our gender test set.
The details of data pre-processing, training, and model architecture can be found in \autoref{app:end-to-end}.

A straightforward way to adapt this vanilla model to produce gender alternatives is to use domain adaptation methods towards the G-Trans dataset (which contains gender structures and alignments). To this end, we train another MT model with the \emph{vanilla bi-text} plus the G-Trans dataset with a prefixed corpus tag \texttt{<gender>} using the loss and serialization described in \autoref{subsec:student}. Adding corpus tags when mixing corpora from different domains has proven to be quite effective \cite{kobus-etal-2017-domain,caswell-etal-2019-tagged,costa2022no}. During inference, this tag is used to decode gender alternatives. We treat this model as the supervised baseline.

Finally, we train a third model, this time augmenting the entire \emph{vanilla bi-text} with gender structures and alignments by passing it through our data augmentation pipeline (using M2M since running GPT at scale is cost-prohibitive).

To measure the impact of our approach on general domain translation performance, we evaluate the models on the FLoRes \cite{costa2022no} test set. Since FLoRes references don't contain gender structures, we also remove gender structures from the outputs of our models (if any are present) while evaluating against FLoRes. We do so by choosing the gender form which is more probable according to the model: concretely, for every gender structure \texttt{BEG} \texttt{\color{brown}M} \texttt{MID} \texttt{\color{teal}F} \texttt{END}, we choose either \texttt{\color{brown}M} or \texttt{\color{teal}F} depending on which phrase has a higher average token log probability.
\subsection{End-to-end MT model results}
\label{subsec:end2end-model-results}
\autoref{tab:student-results} summarizes the results of these models. The vanilla model cannot generate alternatives and shows a huge bias towards generating masculine forms ($\delta$-BLEU ranging from $5.3$ to $12.5$ points). This bias is greatly reduced by the supervised baseline. The model trained on augmented data further reduces the bias and obtains the best performance in terms of alternative metrics, alignment accuracy, and $\delta$-BLEU. This shows the effectiveness of the data augmentation pipeline. Augmented data also allows us to train a competitive system for English-Italian which lacks supervised data.

Results on general domain translation quality (Column FLoRes BLEU from \autoref{tab:student-results}) show that compared to the vanilla baseline, the model trained on augmented data suffers no degradation on English to German and Spanish and some degradations ($-0.3$ to $-0.7$ BLEU) on Engish to French, Portuguese, Russian and Italian.


\section{Conclusion and Future Work}
In this work, we study the task of generating entity-level alternatives when translating a sentence with gender ambiguities into a language with grammatical gender. We open source first train datasets, encouraging future research towards this task, and develop a data augmentation pipeline that leverages pre-trained MT models and LLMs to generate even larger train sets. Finally, we demonstrate that this data can be used effectively to train deployment-friendly MT models that generate alternatives without any additional inference cost or model components.

Our models and pipeline can enable new translation UIs that support fine-grained gender control and can also find applications in aiding human translators to automatically point out ambiguities and recommend alternative translations.

Future work includes exploring other genderless source languages apart from English (e.g., Chinese, Korean, and Japanese) and associated challenges, as well as extending the approach to non-binary and gender-neutral forms \cite{doi:10.1080/0907676X.2023.2268654,piergentili-etal-2023-hi,savoldi-etal-2024-prompt}.
\section*{Bias Statement}
\label{sec:bias-state}
This work focuses on the bias a machine translation system can manifest by solely generating one translation from multiple valid ones that exist with respect to grammatical gender when translating from English to a more gendered language, e.g., French. Singling out one translation as such without offering users the ability to modify the output to match the grammatical gender the user intends for each entity causes two categories of harm: representation harm and quality-of-service harm \cite{10.1145/3313831.3376445,blodgett-etal-2020-language}. It causes representational harm by reflecting the potential stereotypes that lead to the default translation (e.g., between occupations and gender) and quality-of-service harm by failing the users who need the output in the target language to be in a grammatical gender case other than what is generated by default. Our work advocates and proposes a solution for enabling users to choose from all equally correct translation alternatives.

\section*{Limitations}
\label{sec:limits}
All mentions of ``gender'' in this work refer to the grammatical gender present in many languages of the world that are not genderless. Grammatical gender in linguistics is distinct from social gender: while grammatical gender is essentially a noun class system, the discussion surrounding social gender (male, female, nonbinary) encompasses a much more complex set of concepts, e.g., social constructs, norms, roles, and gender identities. Building effective solutions that facilitate inclusive conversations on these topics is not only an open problem in NLP, but many fields.

Moreover, the ambiguities in the linguistic grammatical gender are assumed to be, as in most of the gendered languages, binary: masculine and feminine. However, many languages have more grammatical genders (i.e., noun classes): e.g., Worrorra has masculine, feminine, terrestrial, celestial, and collective.

As such, our proposed resources, as presented so far, fall short of generating entity-level gender-neutral translations or disambiguation beyond the binary system of masculine/feminine. However, it's noteworthy that our pipeline, paired with suitable data resources, e.g., gender-neutral terms for lattice rescoring, forms a powerful instrument for addressing such more challenging settings.

\section*{Acknowledgements}
\label{sec:acks}
We would like to thank Yi-Hsiu Liao, Hendra Setiawan, and Telmo Pessoa Pires for their contributions and discussions through different stages of the project, Matthias Sperber, António Luís Vilarinho dos Santos Lopes, and USC ISI's CUTELABNAME members for their constructive feedback on the paper drafts, and the whole Machine Translation team at Apple for their support for the project.

\bibliography{anthology,custom}

\begin{thebibliography}{45}
\expandafter\ifx\csname natexlab\endcsname\relax\def\natexlab#1{#1}\fi

\bibitem[{Alhafni et~al.(2022)Alhafni, Habash, and Bouamor}]{alhafni-etal-2022-user}
Bashar Alhafni, Nizar Habash, and Houda Bouamor. 2022.
\newblock \href {https://doi.org/10.18653/v1/2022.naacl-main.46} {User-centric gender rewriting}.
\newblock In \emph{Proceedings of the 2022 Conference of the North American Chapter of the Association for Computational Linguistics: Human Language Technologies}, pages 618--631, Seattle, United States. Association for Computational Linguistics.

\bibitem[{Barrault et~al.(2019)Barrault, Bojar, Costa-juss{\`a}, Federmann, Fishel, Graham, Haddow, Huck, Koehn, Malmasi, Monz, M{\"u}ller, Pal, Post, and Zampieri}]{barrault-etal-2019-findings}
Lo{\"\i}c Barrault, Ond{\v{r}}ej Bojar, Marta~R. Costa-juss{\`a}, Christian Federmann, Mark Fishel, Yvette Graham, Barry Haddow, Matthias Huck, Philipp Koehn, Shervin Malmasi, Christof Monz, Mathias M{\"u}ller, Santanu Pal, Matt Post, and Marcos Zampieri. 2019.
\newblock \href {https://doi.org/10.18653/v1/W19-5301} {Findings of the 2019 conference on machine translation ({WMT}19)}.
\newblock In \emph{Proceedings of the Fourth Conference on Machine Translation (Volume 2: Shared Task Papers, Day 1)}, pages 1--61, Florence, Italy. Association for Computational Linguistics.

\bibitem[{Bentivogli et~al.(2020)Bentivogli, Savoldi, Negri, Di~Gangi, Cattoni, and Turchi}]{bentivogli-etal-2020-gender}
Luisa Bentivogli, Beatrice Savoldi, Matteo Negri, Mattia~A. Di~Gangi, Roldano Cattoni, and Marco Turchi. 2020.
\newblock \href {https://doi.org/10.18653/v1/2020.acl-main.619} {Gender in danger? evaluating speech translation technology on the {M}u{ST}-{SHE} corpus}.
\newblock In \emph{Proceedings of the 58th Annual Meeting of the Association for Computational Linguistics}, pages 6923--6933, Online. Association for Computational Linguistics.

\bibitem[{Blodgett et~al.(2020)Blodgett, Barocas, Daum{\'e}~III, and Wallach}]{blodgett-etal-2020-language}
Su~Lin Blodgett, Solon Barocas, Hal Daum{\'e}~III, and Hanna Wallach. 2020.
\newblock \href {https://doi.org/10.18653/v1/2020.acl-main.485} {Language (technology) is power: A critical survey of {``}bias{''} in {NLP}}.
\newblock In \emph{Proceedings of the 58th Annual Meeting of the Association for Computational Linguistics}, pages 5454--5476, Online. Association for Computational Linguistics.

\bibitem[{Brown et~al.(2020)Brown, Mann, Ryder, Subbiah, Kaplan, Dhariwal, Neelakantan, Shyam, Sastry, Askell, Agarwal, Herbert-Voss, Krueger, Henighan, Child, Ramesh, Ziegler, Wu, Winter, Hesse, Chen, Sigler, Litwin, Gray, Chess, Clark, Berner, McCandlish, Radford, Sutskever, and Amodei}]{NEURIPS2020_1457c0d6}
Tom Brown, Benjamin Mann, Nick Ryder, Melanie Subbiah, Jared~D Kaplan, Prafulla Dhariwal, Arvind Neelakantan, Pranav Shyam, Girish Sastry, Amanda Askell, Sandhini Agarwal, Ariel Herbert-Voss, Gretchen Krueger, Tom Henighan, Rewon Child, Aditya Ramesh, Daniel Ziegler, Jeffrey Wu, Clemens Winter, Chris Hesse, Mark Chen, Eric Sigler, Mateusz Litwin, Scott Gray, Benjamin Chess, Jack Clark, Christopher Berner, Sam McCandlish, Alec Radford, Ilya Sutskever, and Dario Amodei. 2020.
\newblock \href {https://proceedings.neurips.cc/paper_files/paper/2020/file/1457c0d6bfcb4967418bfb8ac142f64a-Paper.pdf} {Language models are few-shot learners}.
\newblock In \emph{Advances in Neural Information Processing Systems}, volume~33, pages 1877--1901. Curran Associates, Inc.

\bibitem[{Caswell et~al.(2019)Caswell, Chelba, and Grangier}]{caswell-etal-2019-tagged}
Isaac Caswell, Ciprian Chelba, and David Grangier. 2019.
\newblock \href {https://doi.org/10.18653/v1/W19-5206} {Tagged back-translation}.
\newblock In \emph{Proceedings of the Fourth Conference on Machine Translation (Volume 1: Research Papers)}, pages 53--63, Florence, Italy. Association for Computational Linguistics.

\bibitem[{Chen and Eisele(2012)}]{chen-eisele-2012-multiun}
Yu~Chen and Andreas Eisele. 2012.
\newblock \href {http://www.lrec-conf.org/proceedings/lrec2012/pdf/641_Paper.pdf} {{M}ulti{UN} v2: {UN} documents with multilingual alignments}.
\newblock In \emph{Proceedings of the Eighth International Conference on Language Resources and Evaluation ({LREC}'12)}, pages 2500--2504, Istanbul, Turkey. European Language Resources Association (ELRA).

\bibitem[{Conneau et~al.(2020)Conneau, Khandelwal, Goyal, Chaudhary, Wenzek, Guzm{\'a}n, Grave, Ott, Zettlemoyer, and Stoyanov}]{conneau-etal-2020-unsupervised}
Alexis Conneau, Kartikay Khandelwal, Naman Goyal, Vishrav Chaudhary, Guillaume Wenzek, Francisco Guzm{\'a}n, Edouard Grave, Myle Ott, Luke Zettlemoyer, and Veselin Stoyanov. 2020.
\newblock \href {https://doi.org/10.18653/v1/2020.acl-main.747} {Unsupervised cross-lingual representation learning at scale}.
\newblock In \emph{Proceedings of the 58th Annual Meeting of the Association for Computational Linguistics}, pages 8440--8451, Online. Association for Computational Linguistics.

\bibitem[{Costa{-}juss{\`{a}} et~al.(2022)Costa{-}juss{\`{a}}, Cross, {\c{C}}elebi, Elbayad, Heafield, Heffernan, Kalbassi, Lam, Licht, Maillard, Sun, Wang, Wenzek, Youngblood, Akula, Barrault, Gonzalez, Hansanti, Hoffman, Jarrett, Sadagopan, Rowe, Spruit, Tran, Andrews, Ayan, Bhosale, Edunov, Fan, Gao, Goswami, Guzm{\'{a}}n, Koehn, Mourachko, Ropers, Saleem, Schwenk, and Wang}]{costa2022no}
Marta~R. Costa{-}juss{\`{a}}, James Cross, Onur {\c{C}}elebi, Maha Elbayad, Kenneth Heafield, Kevin Heffernan, Elahe Kalbassi, Janice Lam, Daniel Licht, Jean Maillard, Anna Sun, Skyler Wang, Guillaume Wenzek, Al~Youngblood, Bapi Akula, Lo{\"{\i}}c Barrault, Gabriel~Mejia Gonzalez, Prangthip Hansanti, John Hoffman, Semarley Jarrett, Kaushik~Ram Sadagopan, Dirk Rowe, Shannon Spruit, Chau Tran, Pierre Andrews, Necip~Fazil Ayan, Shruti Bhosale, Sergey Edunov, Angela Fan, Cynthia Gao, Vedanuj Goswami, Francisco Guzm{\'{a}}n, Philipp Koehn, Alexandre Mourachko, Christophe Ropers, Safiyyah Saleem, Holger Schwenk, and Jeff Wang. 2022.
\newblock \href {https://doi.org/10.48550/ARXIV.2207.04672} {No language left behind: Scaling human-centered machine translation}.
\newblock \emph{CoRR}, abs/2207.04672.

\bibitem[{Costa-jussà et~al.(2022)Costa-jussà, Escolano, Basta, Ferrando, Batlle, and Kharitonova}]{Costa-jussà_Escolano_Basta_Ferrando_Batlle_Kharitonova_2022}
Marta~R. Costa-jussà, Carlos Escolano, Christine Basta, Javier Ferrando, Roser Batlle, and Ksenia Kharitonova. 2022.
\newblock \href {https://doi.org/10.1609/aaai.v36i11.21442} {Interpreting gender bias in neural machine translation: Multilingual architecture matters}.
\newblock \emph{Proceedings of the AAAI Conference on Artificial Intelligence}, 36(11):11855--11863.

\bibitem[{Currey et~al.(2022)Currey, Nadejde, Pappagari, Mayer, Lauly, Niu, Hsu, and Dinu}]{currey-etal-2022-mt}
Anna Currey, Maria Nadejde, Raghavendra~Reddy Pappagari, Mia Mayer, Stanislas Lauly, Xing Niu, Benjamin Hsu, and Georgiana Dinu. 2022.
\newblock \href {https://doi.org/10.18653/v1/2022.emnlp-main.288} {{MT}-{G}en{E}val: A counterfactual and contextual dataset for evaluating gender accuracy in machine translation}.
\newblock In \emph{Proceedings of the 2022 Conference on Empirical Methods in Natural Language Processing}, pages 4287--4299, Abu Dhabi, United Arab Emirates. Association for Computational Linguistics.

\bibitem[{Fan et~al.(2021)Fan, Bhosale, Schwenk, Ma, El-Kishky, Goyal, Baines, Celebi, Wenzek, Chaudhary, Goyal, Birch, Liptchinsky, Edunov, Grave, Auli, and Joulin}]{10.5555/3546258.3546365}
Angela Fan, Shruti Bhosale, Holger Schwenk, Zhiyi Ma, Ahmed El-Kishky, Siddharth Goyal, Mandeep Baines, Onur Celebi, Guillaume Wenzek, Vishrav Chaudhary, Naman Goyal, Tom Birch, Vitaliy Liptchinsky, Sergey Edunov, Edouard Grave, Michael Auli, and Armand Joulin. 2021.
\newblock Beyond english-centric multilingual machine translation.
\newblock \emph{J. Mach. Learn. Res.}, 22(1).

\bibitem[{Garg et~al.(2019)Garg, Peitz, Nallasamy, and Paulik}]{garg-etal-2019-jointly}
Sarthak Garg, Stephan Peitz, Udhyakumar Nallasamy, and Matthias Paulik. 2019.
\newblock \href {https://doi.org/10.18653/v1/D19-1453} {Jointly learning to align and translate with transformer models}.
\newblock In \emph{Proceedings of the 2019 Conference on Empirical Methods in Natural Language Processing and the 9th International Joint Conference on Natural Language Processing (EMNLP-IJCNLP)}, pages 4453--4462, Hong Kong, China. Association for Computational Linguistics.

\bibitem[{Geng and Liu(2023)}]{openlm2023openllama}
Xinyang Geng and Hao Liu. 2023.
\newblock \href {https://github.com/openlm-research/open_llama} {Openllama: An open reproduction of llama}.

\bibitem[{Gowda et~al.(2021)Gowda, Zhang, Mattmann, and May}]{gowda-etal-2021-many}
Thamme Gowda, Zhao Zhang, Chris Mattmann, and Jonathan May. 2021.
\newblock \href {https://doi.org/10.18653/v1/2021.acl-demo.37} {Many-to-{E}nglish machine translation tools, data, and pretrained models}.
\newblock In \emph{Proceedings of the 59th Annual Meeting of the Association for Computational Linguistics and the 11th International Joint Conference on Natural Language Processing: System Demonstrations}, pages 306--316, Online. Association for Computational Linguistics.

\bibitem[{Habash et~al.(2019)Habash, Bouamor, and Chung}]{habash-etal-2019-automatic}
Nizar Habash, Houda Bouamor, and Christine Chung. 2019.
\newblock \href {https://doi.org/10.18653/v1/W19-3822} {Automatic gender identification and reinflection in {A}rabic}.
\newblock In \emph{Proceedings of the First Workshop on Gender Bias in Natural Language Processing}, pages 155--165, Florence, Italy. Association for Computational Linguistics.

\bibitem[{Johnson(2020)}]{johnson2020}
Melvin Johnson. 2020.
\newblock A scalable approach to reducing gender bias in google translate.
\newblock \url{https://blog.research.google/2020/04/a-scalable-approach-to-reducing-gender.html}.
\newblock Accessed: 2024-01-28.

\bibitem[{Johnson et~al.(2017)Johnson, Schuster, Le, Krikun, Wu, Chen, Thorat, Vi{\'e}gas, Wattenberg, Corrado, Hughes, and Dean}]{johnson-etal-2017-googles}
Melvin Johnson, Mike Schuster, Quoc~V. Le, Maxim Krikun, Yonghui Wu, Zhifeng Chen, Nikhil Thorat, Fernanda Vi{\'e}gas, Martin Wattenberg, Greg Corrado, Macduff Hughes, and Jeffrey Dean. 2017.
\newblock \href {https://doi.org/10.1162/tacl_a_00065} {{G}oogle{'}s multilingual neural machine translation system: Enabling zero-shot translation}.
\newblock \emph{Transactions of the Association for Computational Linguistics}, 5:339--351.

\bibitem[{Kobus et~al.(2017)Kobus, Crego, and Senellart}]{kobus-etal-2017-domain}
Catherine Kobus, Josep Crego, and Jean Senellart. 2017.
\newblock \href {https://doi.org/10.26615/978-954-452-049-6_049} {Domain control for neural machine translation}.
\newblock In \emph{Proceedings of the International Conference Recent Advances in Natural Language Processing, {RANLP} 2017}, pages 372--378, Varna, Bulgaria. INCOMA Ltd.

\bibitem[{Koehn(2005)}]{koehn-2005-europarl}
Philipp Koehn. 2005.
\newblock \href {https://aclanthology.org/2005.mtsummit-papers.11} {{E}uroparl: A parallel corpus for statistical machine translation}.
\newblock In \emph{Proceedings of Machine Translation Summit X: Papers}, pages 79--86, Phuket, Thailand.

\bibitem[{Kuczmarski and Johnson(2018)}]{Kuczmarski2018GenderAwareNL}
James Kuczmarski and Melvin Johnson. 2018.
\newblock \href {https://api.semanticscholar.org/CorpusID:198676070} {Gender-aware natural language translation}.

\bibitem[{Kudo(2018)}]{kudo-2018-subword}
Taku Kudo. 2018.
\newblock \href {https://doi.org/10.18653/v1/P18-1007} {Subword regularization: Improving neural network translation models with multiple subword candidates}.
\newblock In \emph{Proceedings of the 56th Annual Meeting of the Association for Computational Linguistics (Volume 1: Long Papers)}, pages 66--75, Melbourne, Australia. Association for Computational Linguistics.

\bibitem[{Lardelli(2023)}]{doi:10.1080/0907676X.2023.2268654}
Manuel Lardelli. 2023.
\newblock \href {https://doi.org/10.1080/0907676X.2023.2268654} {Gender-fair translation: a case study beyond the binary}.
\newblock \emph{Perspectives}, 0(0):1--17.

\bibitem[{Madaio et~al.(2020)Madaio, Stark, Wortman~Vaughan, and Wallach}]{10.1145/3313831.3376445}
Michael~A. Madaio, Luke Stark, Jennifer Wortman~Vaughan, and Hanna Wallach. 2020.
\newblock \href {https://doi.org/10.1145/3313831.3376445} {Co-designing checklists to understand organizational challenges and opportunities around fairness in ai}.
\newblock In \emph{Proceedings of the 2020 CHI Conference on Human Factors in Computing Systems}, CHI '20, page 1–14, New York, NY, USA. Association for Computing Machinery.

\bibitem[{Ott et~al.(2019)Ott, Edunov, Baevski, Fan, Gross, Ng, Grangier, and Auli}]{ott-etal-2019-fairseq}
Myle Ott, Sergey Edunov, Alexei Baevski, Angela Fan, Sam Gross, Nathan Ng, David Grangier, and Michael Auli. 2019.
\newblock \href {https://doi.org/10.18653/v1/N19-4009} {fairseq: A fast, extensible toolkit for sequence modeling}.
\newblock In \emph{Proceedings of the 2019 Conference of the North {A}merican Chapter of the Association for Computational Linguistics (Demonstrations)}, pages 48--53, Minneapolis, Minnesota. Association for Computational Linguistics.

\bibitem[{Piergentili et~al.(2023{\natexlab{a}})Piergentili, Fucci, Savoldi, Bentivogli, and Negri}]{piergentili-etal-2023-gender}
Andrea Piergentili, Dennis Fucci, Beatrice Savoldi, Luisa Bentivogli, and Matteo Negri. 2023{\natexlab{a}}.
\newblock \href {https://aclanthology.org/2023.gitt-1.7} {Gender neutralization for an inclusive machine translation: from theoretical foundations to open challenges}.
\newblock In \emph{Proceedings of the First Workshop on Gender-Inclusive Translation Technologies}, pages 71--83, Tampere, Finland. European Association for Machine Translation.

\bibitem[{Piergentili et~al.(2023{\natexlab{b}})Piergentili, Savoldi, Fucci, Negri, and Bentivogli}]{piergentili-etal-2023-hi}
Andrea Piergentili, Beatrice Savoldi, Dennis Fucci, Matteo Negri, and Luisa Bentivogli. 2023{\natexlab{b}}.
\newblock \href {https://doi.org/10.18653/v1/2023.emnlp-main.873} {Hi guys or hi folks? benchmarking gender-neutral machine translation with the {G}e{NTE} corpus}.
\newblock In \emph{Proceedings of the 2023 Conference on Empirical Methods in Natural Language Processing}, pages 14124--14140, Singapore. Association for Computational Linguistics.

\bibitem[{Post(2018)}]{post-2018-call}
Matt Post. 2018.
\newblock \href {https://doi.org/10.18653/v1/W18-6319} {A call for clarity in reporting {BLEU} scores}.
\newblock In \emph{Proceedings of the Third Conference on Machine Translation: Research Papers}, pages 186--191, Brussels, Belgium. Association for Computational Linguistics.

\bibitem[{Rarrick et~al.(2023)Rarrick, Naik, Mathur, Poudel, and Chowdhary}]{10.1145/3600211.3604675}
Spencer Rarrick, Ranjita Naik, Varun Mathur, Sundar Poudel, and Vishal Chowdhary. 2023.
\newblock \href {https://doi.org/10.1145/3600211.3604675} {Gate: A challenge set for gender-ambiguous translation examples}.
\newblock In \emph{Proceedings of the 2023 AAAI/ACM Conference on AI, Ethics, and Society}, AIES '23, page 845–854, New York, NY, USA. Association for Computing Machinery.

\bibitem[{Renduchintala et~al.(2021)Renduchintala, Diaz, Heafield, Li, and Diab}]{renduchintala-etal-2021-gender}
Adithya Renduchintala, Denise Diaz, Kenneth Heafield, Xian Li, and Mona Diab. 2021.
\newblock \href {https://doi.org/10.18653/v1/2021.acl-short.15} {Gender bias amplification during speed-quality optimization in neural machine translation}.
\newblock In \emph{Proceedings of the 59th Annual Meeting of the Association for Computational Linguistics and the 11th International Joint Conference on Natural Language Processing (Volume 2: Short Papers)}, pages 99--109, Online. Association for Computational Linguistics.

\bibitem[{Rozis and Skadi{\c{n}}{\v{s}}(2017)}]{rozis-skadins-2017-tilde}
Roberts Rozis and Raivis Skadi{\c{n}}{\v{s}}. 2017.
\newblock \href {https://aclanthology.org/W17-0235} {Tilde {MODEL} - multilingual open data for {EU} languages}.
\newblock In \emph{Proceedings of the 21st Nordic Conference on Computational Linguistics}, pages 263--265, Gothenburg, Sweden. Association for Computational Linguistics.

\bibitem[{S{\'{a}}nchez et~al.(2023)S{\'{a}}nchez, Andrews, Stenetorp, Artetxe, and Costa{-}juss{\`{a}}}]{DBLP:journals/corr/abs-2309-03175}
Eduardo S{\'{a}}nchez, Pierre Andrews, Pontus Stenetorp, Mikel Artetxe, and Marta~R. Costa{-}juss{\`{a}}. 2023.
\newblock \href {https://doi.org/10.48550/ARXIV.2309.03175} {Gender-specific machine translation with large language models}.
\newblock \emph{CoRR}, abs/2309.03175.

\bibitem[{Saunders and Byrne(2020)}]{saunders-byrne-2020-reducing}
Danielle Saunders and Bill Byrne. 2020.
\newblock \href {https://doi.org/10.18653/v1/2020.acl-main.690} {Reducing gender bias in neural machine translation as a domain adaptation problem}.
\newblock In \emph{Proceedings of the 58th Annual Meeting of the Association for Computational Linguistics}, pages 7724--7736, Online. Association for Computational Linguistics.

\bibitem[{Saunders et~al.(2022)Saunders, Sallis, and Byrne}]{saunders-etal-2022-first}
Danielle Saunders, Rosie Sallis, and Bill Byrne. 2022.
\newblock \href {https://doi.org/10.18653/v1/2022.findings-acl.301} {First the worst: Finding better gender translations during beam search}.
\newblock In \emph{Findings of the Association for Computational Linguistics: ACL 2022}, pages 3814--3823, Dublin, Ireland. Association for Computational Linguistics.

\bibitem[{Savoldi et~al.(2024)Savoldi, Piergentili, Fucci, Negri, and Bentivogli}]{savoldi-etal-2024-prompt}
Beatrice Savoldi, Andrea Piergentili, Dennis Fucci, Matteo Negri, and Luisa Bentivogli. 2024.
\newblock \href {https://aclanthology.org/2024.eacl-short.23} {A prompt response to the demand for automatic gender-neutral translation}.
\newblock In \emph{Proceedings of the 18th Conference of the European Chapter of the Association for Computational Linguistics (Volume 2: Short Papers)}, pages 256--267, St. Julian{'}s, Malta. Association for Computational Linguistics.

\bibitem[{Schwenk et~al.(2021)Schwenk, Chaudhary, Sun, Gong, and Guzm{\'a}n}]{schwenk-etal-2021-wikimatrix}
Holger Schwenk, Vishrav Chaudhary, Shuo Sun, Hongyu Gong, and Francisco Guzm{\'a}n. 2021.
\newblock \href {https://doi.org/10.18653/v1/2021.eacl-main.115} {{W}iki{M}atrix: Mining 135{M} parallel sentences in 1620 language pairs from {W}ikipedia}.
\newblock In \emph{Proceedings of the 16th Conference of the European Chapter of the Association for Computational Linguistics: Main Volume}, pages 1351--1361, Online. Association for Computational Linguistics.

\bibitem[{Stafanovi{\v{c}}s et~al.(2020)Stafanovi{\v{c}}s, Bergmanis, and Pinnis}]{stafanovics-etal-2020-mitigating}
Art{\=u}rs Stafanovi{\v{c}}s, Toms Bergmanis, and M{\=a}rcis Pinnis. 2020.
\newblock \href {https://aclanthology.org/2020.wmt-1.73} {Mitigating gender bias in machine translation with target gender annotations}.
\newblock In \emph{Proceedings of the Fifth Conference on Machine Translation}, pages 629--638, Online. Association for Computational Linguistics.

\bibitem[{Stanovsky et~al.(2019)Stanovsky, Smith, and Zettlemoyer}]{stanovsky-etal-2019-evaluating}
Gabriel Stanovsky, Noah~A. Smith, and Luke Zettlemoyer. 2019.
\newblock \href {https://doi.org/10.18653/v1/P19-1164} {Evaluating gender bias in machine translation}.
\newblock In \emph{Proceedings of the 57th Annual Meeting of the Association for Computational Linguistics}, pages 1679--1684, Florence, Italy. Association for Computational Linguistics.

\bibitem[{Sun et~al.(2019)Sun, Gaut, Tang, Huang, ElSherief, Zhao, Mirza, Belding, Chang, and Wang}]{sun-etal-2019-mitigating}
Tony Sun, Andrew Gaut, Shirlyn Tang, Yuxin Huang, Mai ElSherief, Jieyu Zhao, Diba Mirza, Elizabeth Belding, Kai-Wei Chang, and William~Yang Wang. 2019.
\newblock \href {https://doi.org/10.18653/v1/P19-1159} {Mitigating gender bias in natural language processing: Literature review}.
\newblock In \emph{Proceedings of the 57th Annual Meeting of the Association for Computational Linguistics}, pages 1630--1640, Florence, Italy. Association for Computational Linguistics.

\bibitem[{Sun et~al.(2021)Sun, Webster, Shah, Wang, and Johnson}]{DBLP:journals/corr/abs-2102-06788}
Tony Sun, Kellie Webster, Apurva Shah, William~Yang Wang, and Melvin Johnson. 2021.
\newblock \href {http://arxiv.org/abs/2102.06788} {They, them, theirs: Rewriting with gender-neutral english}.
\newblock \emph{CoRR}, abs/2102.06788.

\bibitem[{Tiedemann(2012)}]{TIEDEMANN12.463}
Jörg Tiedemann. 2012.
\newblock Parallel data, tools and interfaces in opus.
\newblock In \emph{Proceedings of the Eight International Conference on Language Resources and Evaluation (LREC'12)}, Istanbul, Turkey. European Language Resources Association (ELRA).

\bibitem[{Touvron et~al.(2023)Touvron, Lavril, Izacard, Martinet, Lachaux, Lacroix, Rozi{\`{e}}re, Goyal, Hambro, Azhar, Rodriguez, Joulin, Grave, and Lample}]{DBLP:journals/corr/abs-2302-13971}
Hugo Touvron, Thibaut Lavril, Gautier Izacard, Xavier Martinet, Marie{-}Anne Lachaux, Timoth{\'{e}}e Lacroix, Baptiste Rozi{\`{e}}re, Naman Goyal, Eric Hambro, Faisal Azhar, Aur{\'{e}}lien Rodriguez, Armand Joulin, Edouard Grave, and Guillaume Lample. 2023.
\newblock \href {https://doi.org/10.48550/ARXIV.2302.13971} {Llama: Open and efficient foundation language models}.
\newblock \emph{CoRR}, abs/2302.13971.

\bibitem[{Vanmassenhove et~al.(2021)Vanmassenhove, Emmery, and Shterionov}]{vanmassenhove-etal-2021-neutral}
Eva Vanmassenhove, Chris Emmery, and Dimitar Shterionov. 2021.
\newblock \href {https://doi.org/10.18653/v1/2021.emnlp-main.704} {{N}eu{T}ral {R}ewriter: {A} rule-based and neural approach to automatic rewriting into gender neutral alternatives}.
\newblock In \emph{Proceedings of the 2021 Conference on Empirical Methods in Natural Language Processing}, pages 8940--8948, Online and Punta Cana, Dominican Republic. Association for Computational Linguistics.

\bibitem[{Wolf et~al.(2020)Wolf, Debut, Sanh, Chaumond, Delangue, Moi, Cistac, Rault, Louf, Funtowicz, Davison, Shleifer, von Platen, Ma, Jernite, Plu, Xu, Le~Scao, Gugger, Drame, Lhoest, and Rush}]{wolf-etal-2020-transformers}
Thomas Wolf, Lysandre Debut, Victor Sanh, Julien Chaumond, Clement Delangue, Anthony Moi, Pierric Cistac, Tim Rault, Remi Louf, Morgan Funtowicz, Joe Davison, Sam Shleifer, Patrick von Platen, Clara Ma, Yacine Jernite, Julien Plu, Canwen Xu, Teven Le~Scao, Sylvain Gugger, Mariama Drame, Quentin Lhoest, and Alexander Rush. 2020.
\newblock \href {https://doi.org/10.18653/v1/2020.emnlp-demos.6} {Transformers: State-of-the-art natural language processing}.
\newblock In \emph{Proceedings of the 2020 Conference on Empirical Methods in Natural Language Processing: System Demonstrations}, pages 38--45, Online. Association for Computational Linguistics.

\bibitem[{Ylonen(2022)}]{ylonen-2022-wiktextract}
Tatu Ylonen. 2022.
\newblock \href {https://aclanthology.org/2022.lrec-1.140} {Wiktextract: {W}iktionary as machine-readable structured data}.
\newblock In \emph{Proceedings of the Thirteenth Language Resources and Evaluation Conference}, pages 1317--1325, Marseille, France. European Language Resources Association.

\end{thebibliography}

\appendix
\newpage 
{\renewcommand{\arraystretch}{1.1}
\begin{table*}
\centering
\small
\begin{tabular}{|c|c|ccccc|}
\hline
Dataset & Statistic & En-De & En-Es & En-Fr & En-Pt & En-Ru \\ \hline\hline
\multirow{4}{*}{G-Tag} & Sentences & $11.7$ & $13.5$ & $13.3$ & $13.3$ & $10.3$ \\ 
& Ambiguous entities & $13.8$ & $14$ & $13.2$ & $14.6$ & $11.3$ \\
& Masculine entities & $7.4$  & $7.8$  & $7.6$  & $7.9$  & $6.6$  \\
& Feminine entities  & $6.1$  & $7$  & $6.7$  & $7$  & $5.6$  \\ \hline
\multirow{3}{*}{G-Trans} & Sentences & $49.4$ & $49.6$ & $49.7$ & $49.6$ & $48.8$ \\
& Ambiguous entities & $69.3$ & $74.7$ & $69.1$ & $73.9$ & $64.1$ \\
& Gender structures  & $77.5$ & $81.7$ & $77.6$ & $83.1$ & $72.5$ \\ \hline
\end{tabular}
\caption{Train set statistics: All numbers are \emph{in thousands}. We sample about $12$k sentences for the G-Tag dataset, roughly containing $2:1:1$ ratio of ambiguous, masculine and feminine entities. About $50$k sentence pairs with ambiguous entities and gender structures are sampled for the G-Trans dataset.}
\label{tab:trainset-statistics}
\end{table*}}
~
\section{Dataset details}
\label{app:dataset-details}
{\renewcommand{\arraystretch}{1.1}
\begin{table}[h]
\centering
\small
\begin{tabular}{|c|c|>{\centering\arraybackslash}p{0.8in}|>{\centering\arraybackslash}p{0.5in}|}
\hline
\multirow{2}{0.5in}{\centering Language Pair} & \multicolumn{3}{c|}{No. of sentences with} \\
   &\multicolumn{1}{c}{Total} & \multicolumn{1}{>{\centering\arraybackslash}p{0.8in}}{1+ Ambiguous entities} & 1+ gender structures \\ \hline\hline
En-De & $3038$ & $2765$ & $2118$ \\
En-Es & $1407$ & $1147$ & $972$ \\
En-Fr & $1564$ & $1292$ & $1006$ \\
En-Pt & $3083$ & $2764$ & $2435$ \\
En-Ru & $3083$ & $2765$ & $1847$ \\ 
En-It & $1312$ & $1018$ & $858$ \\ \hline
\end{tabular}
\caption{Test set statistics: About $80-90\%$ sentences contain at least one gender-ambiguous entity, out of which about $60-80\%$ contain gender structures in the reference.}
\label{tab:testset-statistics}
\end{table}}
\pgfplotsset{width=7cm,compat=1.9}
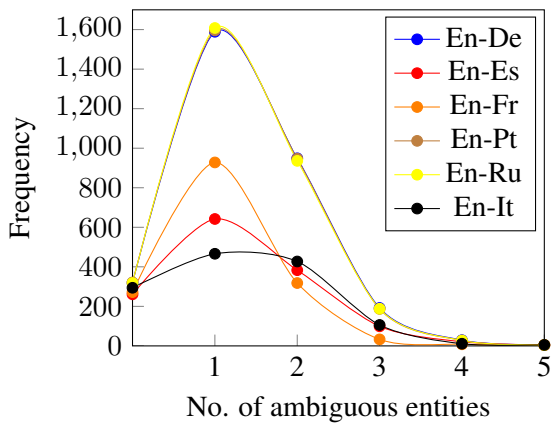
\begin{figure}[h]
\begin{tikzpicture}[trim left=-0.5in]
\begin{axis}[
    xlabel=No. of ambiguous entities,
    ylabel=Frequency,
    xmin=0, xmax=5,
    ymin=0, ymax=1700,
    xtick={1,2,3,4,5},
    xticklabels={1,2,3,4,5},
    xtick pos=bottom,
    ytick={0,200,...,1600},
    ytick pos=left,
            ]
\addplot[smooth,mark=*,blue] plot coordinates {
    (0,319)
    (1,1589)
    (2,949)
    (3,192)
    (4,29)
    (5,4)
};
\addlegendentry{En-De}
\addplot[smooth,mark=*,red] plot coordinates {
    (0, 261)
    (1,642)
    (2,383)
    (3,99)
    (4,21)
    (5,4)
};
\addlegendentry{En-Es}
\addplot[smooth,mark=*,orange] plot coordinates {
    (0, 272)
    (1,928)
    (2,318)
    (3,33)
    (4,8)
    (5,5)
};
\addlegendentry{En-Fr}
\addplot[smooth,mark=*,brown] plot coordinates {
    (0, 321)
    (1,1596)
    (2,947)
    (3,187)
    (4,27)
    (5,4)
};
\addlegendentry{En-Pt}
\addplot[smooth,mark=*,yellow] plot coordinates {
    (0, 318)
    (1,1608)
    (2,936)
    (3,188)
    (4,27)
    (5,4)
};
\addlegendentry{En-Ru}
\addplot[smooth,mark=*,black] plot coordinates {
    (0, 294)
    (1,466)
    (2,427)
    (3,106)
    (4,11)
    (5,5)
};
\addlegendentry{En-It}
\end{axis}
\end{tikzpicture}
\caption{Number of examples v.s. number of ambiguous entities in the test set.}
\label{fig:testset-statistics}
\end{figure}
Detailed train data statistics are listed in Table~\ref{tab:trainset-statistics}. Detailed test set statistics are shown in \autoref{tab:testset-statistics} and \autoref{fig:testset-statistics}.

We had to get the annotations in GATE and MT-GenEval reviewed and post-edited from human annotators because their annotation guidelines differ from ours in the following respects:
\begin{itemize}
\item GATE defines a gender-ambiguous entity as an entity whose gender cannot be inferred from the grammatical sentence context \emph{and} whose gender can influence changes in the translation. This second requirement makes this definition of \emph{ambiguous entity} dependent on the target language/translation. E.g., in ``I am going to the market'', despite the gender of \emph{I} being ambiguous, it would not be marked as an ambiguous entity for English-Spanish, since the Spanish translation does not change based on the gender of \emph{I}. The same entity would be marked as an ambiguous entity in case of English-Hindi where the translation changes based on the gender of \emph{I}.

In our definition of an ambiguous entity, we drop the second requirement, making it independent of the translation and the target language. This enables us to train an ambiguity tagger solely on the Engish source sentences which can be used for any English-X language pair. This, however, forces us to re-annotate the GATE corpus.
\item MT-GenEval corpus contains source sentences with annotated entities whose gender \emph{can} be inferred as masculine/feminine from the sentence context. This provides a valuable test-bed for catching false positive gender alternatives. However, we found that $\sim 50$\% of source sentences also contain one or more ambiguous entities which have not been annotated. Therefore we re-annotate the MT-GenEval corpus as well to mark such entities.
\end{itemize}

Upon the deanonymized publication of this work, we plan to release the datasets under CC BY-SA license.

{\renewcommand{\arraystretch}{1.1}
\begin{table*}
\centering
\small
\begin{tabular}{|>{\centering\arraybackslash}p{0.5in}|cc|cc|cc|}
\hline
\multirow{2}{0.5in}{\centering Language Pair} & \multicolumn{2}{c|}{Ambiguous Entities} & \multicolumn{2}{c|}{Masculine Entities} & \multicolumn{2}{c|}{Feminine Entities}\\
& Precision\% & Recall\% & Precision\% & Recall\% & Precision\% & Recall\% \\\hline \hline
En-De & $93.1$ & $91.4$ & $72.5$ & $83.0$ & $74.7$ & $84.2$ \\
En-Es & $89.8$ & $86.6$ & $70.3$ & $82.3$ & $74.8$ & $83.6$ \\
En-Fr & $90.3$ & $88.1$ & $69.0$ & $80.0$ & $70.0$ & $80.7$ \\
En-Pt & $93.1$ & $91.4$ & $70.6$ & $84.4$ & $73.2$ & $87.8$\\
En-Ru & $93.2$ & $91.3$ & $71.7$ & $83.9$ & $73.6$ & $84.0$\\
En-It & $92.1$ & $89.2$ & $72.3$ & $84.4$ & $72.0$ & $85.7$\\ \hline
\end{tabular}
\caption{Results of tagging different gendered entities by the \texttt{XLM} based tagger.}
\label{tab:tagger-results}
\end{table*}
}
\section{Problem of masculine generics during gender ambiguity annotation}
\label{app:masculine_generics}
It is fairly common to use masculine gendered words to refer to ambiguous entities. In administrative and legal text, masculine gendered words have been used to refer to collection of people \citep{piergentili-etal-2023-gender} for e.g. ``A judge must certify that \textbf{he} has familiarized \textbf{himself} with...''. It is a complex problem to ascertain whether \emph{he} refers to a masculine individual or a group of (ambiguous gendered) people at large. 

In our annotation guidelines we informed the annotators that entities shouldn't be marked as masculine solely because of masculine generic nouns like \emph{actor}, \emph{sportsmen}. However no special guidelines were provided around the trickier case of masculine generic pronouns (\emph{he}, \emph{himself} as shown in the example above)

\section{Synthetically generated train data}
\label{app:train_data}
We used human annotation to collect primary versions of G-Trans and G-Tag datasets (gender train sets) using the annotation process described in \autoref{subsec:test_data}. However, we are unable to release these “human-annotated” sets publicly due to legal and proprietary data restrictions. To make our approach and results reproducible to the community, we instead plan to release "synthetically generated sets" generated as follows: we trained our data augmentation pipeline (described in \autoref{sec:data-aug}) on the “human-annotated” training sets and then ran the data augmentation pipeline on corpora mentioned \autoref{subsec:train_data}. We then sampled the G-Trans and G-Tag datasets from the pipeline results and use them throughout our work.

\section{Gender-ambiguous entity detector}
\label{app:tagger}
The gender-ambiguous entity detector is fine-tuned using the following hyper-parameters:
\begin{itemize}[noitemsep,topsep=1pt,leftmargin=.15in]
\item \texttt{batch size: 64}
\item \texttt{epochs: 2}
\item \texttt{learning rate: 2e-5}
\item \texttt{tokenizer: intl from sacrebleu library}
\item \texttt{subword model: default xlm-roberta-large tokenizer}
\item \texttt{output labels: <A>} (ambiguous), \texttt{<M>} (masculine), \texttt{<F>} (feminine), \texttt{<N>} (not an entity)
\item \texttt{linear tagging layer:} $1024 \times 4$
\item Architecture hyper-parameters can be found by loading \texttt{xlm-roberta-large} using \texttt{AutoModelForTokenClassification} in \texttt{transformers}.
\item The tagging loss is applied only on the first sub-word of each token. The prediction for each token is computed based on the label output for the first sub-word.
\item We fine-tune all the parameters of the pre-trained model along with the added linear layer.
\item All reported results are gathered from a single run.
\end{itemize}
\autoref{tab:tagger-results} summarizes the results of the detector on tagging entities of different genders.

\section{Generating all-masculine/feminine translations by finetuned-M2M model}
\label{app:finetuned-m2m}
We fine-tuned a pre-trained \texttt{M2M-1.2B} model with the following hyper-parameters:
\begin{itemize}[noitemsep,topsep=1pt,leftmargin=.15in]
\item \texttt{batch size: 8192}
\item \texttt{learning rate: 3e-5}
\item \texttt{encoder layerdrop: disabled}
\item \texttt{decoder layerdrop: disabled}
\item Rest of the hyper-parameters are the same as the pre-trained model.
\item We fine-tune for a total of \texttt{40000} steps and select the best checkpoint based on loss on a held out validation set.
\item We use the sub-word model and dictionaries of the pre-trained M2M model. However, we add gender assignment tags (\texttt{<M>} and \texttt{<F>}) as new entries in the dictionary and train their embeddings from scratch. 
\item We use a beam size of \texttt{5} while decoding all-masculine/feminine translations using lattice-rescoring.
\item All reported results are gathered from a single run.
\end{itemize}
\section{Ablation studies on generating using LLMs}
\label{app:llm-ablation}
We study the effect of three factors on the effectiveness of LLMs for generating all-masculine/feminine translations as part of our data augmentation process: number of in-context examples, prompt design, and choice of LLM.

\begin{table*}[!ht]
\centering
\small
\begin{tabular}{|c|cc|cc|cc|}
\hline
\multirow{2}{1.5cm}{\centering Language Pair} & \multirow{2}{*}{LLM} & \multirow{2}{*}{Prompting View} & \multicolumn{2}{c|}{Alternatives Metrics $\uparrow$} & \multicolumn{2}{c|}{Structure Metrics$\uparrow$} \\
& & & Precision\% & Recall\% & Precision\% & Recall\% \\ \hline\hline
\multirow{4}{*}{En–De} & \multirow{2}{*}{GPT} & Generator & $\mathbf{91.5}$ & $81.8$ & $73.2$ & $74.8$ \\
      & & Editor & $89.4$ & $\mathbf{86.1}$ & $\mathbf{73.9}$ & $\mathbf{76}$ \\
      \cline{2-7}
      & \multirow{2}{*}{OpenLLaMA} & Generator & $91.5$ & $26.6$ & $\mathbf{48.2}$ & $\mathbf{41.4}$ \\
      & & Editor & $\mathbf{92.5}$ & $\mathbf{47.8}$ & $43.4$ & $37.6$ \\ \hline\hline
\multirow{4}{*}{En–Es} & \multirow{2}{*}{GPT} & Generator & $90.3$ & $87.9$ & $60.4$ & $66.4$ \\
      & & Editor & $\mathbf{91.6}$ & $\mathbf{92.4}$ & $\mathbf{63.5}$ & $\mathbf{69.5}$ \\
      \cline{2-7}
      & \multirow{2}{*}{OpenLLaMA} & Generator & $67.5$ & $7.9$ & $31.1$ & $26.9$ \\
      & & Editor & $\mathbf{91.4}$ & $\mathbf{34}$ & $\mathbf{52.9}$ & $\mathbf{40.7}$ \\ \hline\hline
\multirow{4}{*}{En–Fr} & \multirow{2}{*}{GPT} & Generator & $87.4$ & $82.3$ & $\mathbf{69.4}$ & $\mathbf{77}$ \\
      & & Editor & $\mathbf{88.1}$ & $\mathbf{86.8}$ & $63.8$ & $75.7$ \\
      \cline{2-7}
      & \multirow{2}{*}{OpenLLaMA} & Generator & $54.6$ & $5.3$ & $24.7$ & $28$ \\
      & & Editor & $\mathbf{85.9}$ & $\mathbf{32.8}$ & $\mathbf{58.4}$ & $\mathbf{52.3}$ \\ \hline\hline
\multirow{4}{*}{En–Pt} & \multirow{2}{*}{GPT} & Generator & $\mathbf{94}$ & $78.1$ & $\mathbf{66}$ & $\mathbf{66.8}$ \\
      & & Editor & $92.8$ & $\mathbf{79.8}$ & $63.3$ & $66.6$ \\
      \cline{2-7}
      & \multirow{2}{*}{OpenLLaMA} & Generator & $89.7$ & $11.8$ & $46.6$ & $32$ \\
      & & Editor & $\mathbf{93.7}$ & $\mathbf{44.8}$ & $\mathbf{54}$ & $\mathbf{43.6}$ \\ \hline\hline
\multirow{4}{*}{En–Ru} & \multirow{2}{*}{GPT} & Generator & $\mathbf{83.9}$ & $\mathbf{61.8}$ & $45.9$ & $45.1$ \\
      & & Editor & $80.1$ & $55.3$ & $\mathbf{48.8}$ & $\mathbf{49.4}$ \\
      \cline{2-7}
      & \multirow{2}{*}{OpenLLaMA} & Generator & $67.6$ & $6.4$ & $8.9$ & $8.4$ \\
      & & Editor & $\mathbf{79.1}$ & $\mathbf{12.1}$ & $\mathbf{27.4}$ & $\mathbf{21.4}$ \\ \hline
\end{tabular}
\caption{LLM Ablation Results.}
\label{tab:llm-ablation}
\end{table*}

\subsection{Number of in-context examples}
In our preliminary experiments, we found using at least four in-context examples to be necessary for our task, with performance starting to plateau thereafter (see the chart below in Figure~\ref{fig:in-context-examples-ablation}). We use six in-context examples in the rest of the experiments.

\pgfplotsset{width=6cm,compat=1.9}
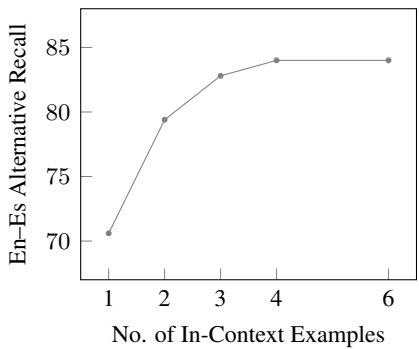
\begin{figure}[h]
\centering
\begin{tikzpicture}[trim left=-0.5in]
\tikzstyle{every node}=[font=\small]
\begin{axis}[
    xlabel=No. of In-Context Examples,
    ylabel=En–Es Alternative Recall,
    xmin=0.5, xmax=6.5,
    ymin=67, ymax=88,
    xtick={1,2,3,4,6},
    xticklabels={1,2,3,4,6},
    xtick pos=bottom,
    ytick={70, 75, 80, 85},
    ytick pos=left,
            ]
\addplot[mark=*,gray,mark size=0.9pt] plot coordinates {
    (1,70.6)
    (2,79.4)
    (3,82.8)
    (4,84)
    (6,84)
};
\end{axis}
\end{tikzpicture}
\caption{Ablation on the number of in-context examples. We use the GPT's alternative recall on English–Spanish as an exemplar. Per this results, we use six in-context examples for prompting.}
\label{fig:in-context-examples-ablation}
\end{figure}

\subsection{Choice of LLM and prompt design}
In addition to GPT (\texttt{gpt-3.5-turbo}), we also experiment with OpenLLaMA (\texttt{OpenLlama-v2-7B}) \cite{openlm2023openllama}, an open reproduction of LLaMA \cite{DBLP:journals/corr/abs-2302-13971}. We find these two to vary in overall performance and robustness to different kinds of prompts.

\begin{figure}[h]
    \centering
    \includegraphics[width=\columnwidth]{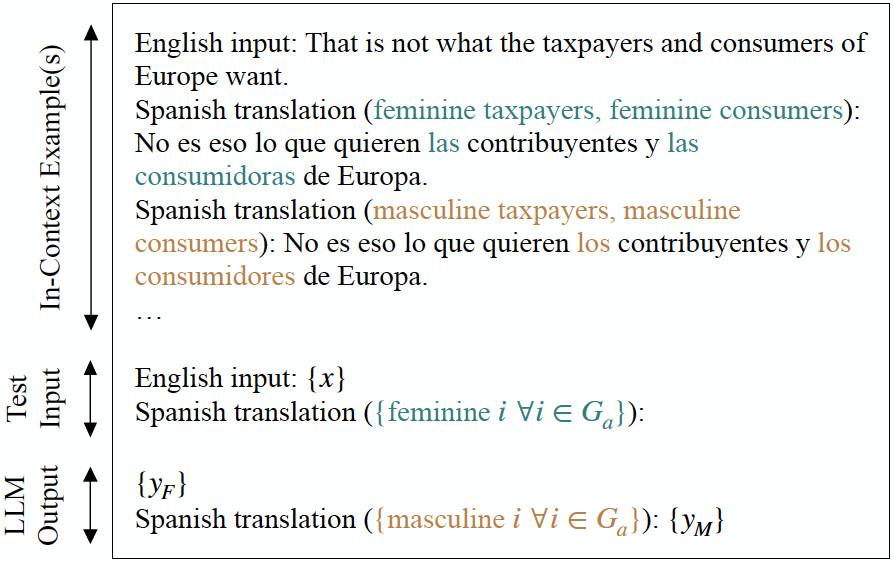}
    \caption{Prompting LLMs using in-context examples to generate translations with all-masculine and all-feminine gender assignments from scratch.}
    \label{fig:icl-generator}
\end{figure}

Specifically, besides the prompt design discussed in the main text, which has the LLM \textit{edit} an existing translation to satisfy the provided grammatical gender requirements, we also experiment with an additional design: given the input and the grammatical gender requirements, we have the LLM generate the translation from scratch (Figure~\ref{fig:icl-generator}). We call the former the editor-view prompting, and the latter the generator-view prompting.

In editor-view prompting, the base translation can be sourced in any number of ways, including using the reference translation, as we did in \autoref{subsec:data-aug-results}. However, to make the study between editor-view and generator-view fair and make sure reference translations do not give any advantage to the editor-view, we first prompt the LLM for base translations (first call) and then have it edit those (second call). This effectively breaks the task of generating gender alternatives down to two separate tasks for LLMs: translation, and then editing.

Table~\ref{tab:llm-ablation} reports and compares the results of prompting each of the two LLMs we experiment with, using each of the two prompt designs we use. All reported results are gathered from a single run. GPT, expectedly, outperforms OpenLLaMA. And while both generally benefit from breaking down the task under the editor-view (and perform better under editor-view than under generator-view), OpenLLaMA conspicuously profits more. Specifically, OpenLLaMA's alternative recall under the generator-view suggests that it fails to generate alternatives following the in-context examples. However, under the editor-view, it is able to follow the in-context examples more. The wider gap between the performance of OpenLLaMA under the two prompting approaches compared to that of GPT, shows that for our task, it's far less robust to different prompt designs.
\section{Aligning gender-ambiguous entities}
\label{app:aligner}
\begin{figure*}
    \centering
    \includegraphics[width=0.75\linewidth]{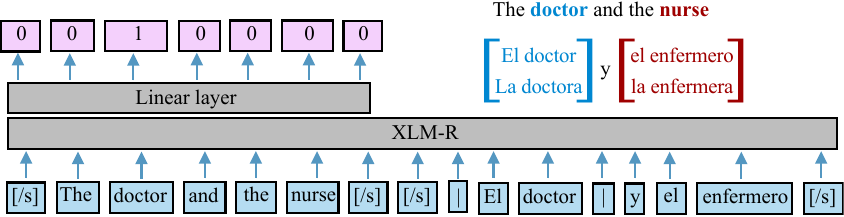}
    \caption{This figure shows an example of aligning the gender structure $\big(\begin{smallmatrix} \text{\color{brown}El doctor}\\ \text{\color{teal}La doctora} \end{smallmatrix}\big)$. The model is fine-tuned to classify the source tokens as being aligned ($1$) or not-aligned ($0$) to this gender structure.}
    \label{fig:alignment}
\end{figure*}
We fine-tune an \texttt{xlm-roberta-large} model for aligning gender structures to their corresponding ambiguous entities using the following hyper-parameters:
\begin{itemize}[noitemsep,topsep=1pt,leftmargin=.15in]
\item \texttt{epochs: 1}
\item \texttt{output labels: 1}(aligned), \texttt{2} (not-aligned)
\item \texttt{linear tagging layer:} $1024 \times 2$
\item Rest of the hyper-parameters are same as the gender-ambiguous entity detector (\autoref{app:tagger}).
\item All reported results are gathered from a single run.
\end{itemize}
\autoref{fig:alignment} shows an example of input and output when aligning a gender structure.
\section{Running data augmentation pipeline on outputs of M2M and GPT}
\label{app:data-augmentation}
In this work we focus on running the data augmentation pipeline over parallel corpora to enrich them with gender structures and gender alignments. However, the pipeline can also be run over \emph{any translation} system to generate entity-level gender alternatives. \autoref{tab:non-oracle-results} shows the results when the data augmentation pipeline is run over translations from the pre-trained M2M and GPT models. 

The pipeline uses fine-tuned M2M when run over translations from the M2M model and the editor-view prompting using GPT when run over translations from GPT. We can see that both M2M and GPT have large bias towards producing masculine translations ($\delta$-BLEU values ranging from $6.5$ to $12.7$ points). The data augmentation pipeline has multiple components and much higher inference cost than the end-end student model, but can produce higher quality gender alternatives when compared to the end-end model (\autoref{tab:student-results} vs. \autoref{tab:non-oracle-results}).

{\renewcommand{\arraystretch}{1.1}
\begin{table*}
\centering
\small
\begin{tabular}{|c|c|cc|ccc|cc|}
\hline
\multirow{2}{1.5cm}{\centering Language Pair} & Model & \multicolumn{2}{c|}{Alternatives Metrics$\uparrow$} & \multicolumn{3}{c|}{BLEU} & \multicolumn{2}{c|}{Structure Metrics$\uparrow$}\\
& & Precision\% & Recall\% & Masc.$\uparrow$ & Fem.$\uparrow$ & $\delta\downarrow$ & Precision\% & Recall\%\\ \hline\hline
\multirow{4}{*}{En–De} & M2M & - & - & $46.8$ & $36.6$ & $10.2$ & - & - \\
      & + Data Augmentation & $\mathbf{92.5}$ & $82.8$ & $46.9$ & $45.7$ & $1.2$ & $64.7$ & $64.2$ \\ \cline{2-9}
      & GPT & - & - & $\mathbf{53.8}$ & $41.4$ & $12.4$ & - & - \\
      & + Data Augmentation & $89.4$ & $\mathbf{86.1}$ & $\mathbf{53.8}$ & $\mathbf{52.7}$ & $\mathbf{1.1}$ & $\mathbf{73.9}$ & $\mathbf{76}$ \\ \hline\hline
\multirow{4}{*}{En–Es} & M2M & - & - & $47.3$ & $37$ & $10.3$ & - & - \\
      & + Data Augmentation & $\mathbf{95.8}$ & $91.3$ & $47.5$ & $46.5$ & $\mathbf{1}$ & $63.2$ & $64$ \\ \cline{2-9}
      & GPT & - & - & $\mathbf{51.8}$ & $40.4$ & $11.4$ & - & - \\
      & + Data Augmentation & $91.6$ & $\mathbf{92.4}$ & $51.5$ & $\mathbf{50.4}$ & $1.1$ & $\mathbf{63.5}$ & $\mathbf{69.5}$ \\\hline\hline
\multirow{4}{*}{En–Fr} & M2M & - & - & $50$ & $41.5$ & $8.5$ & - & - \\
      & + Data Augmentation & $\mathbf{90.7}$ & $84$ & $52.4$ & $48.8$ & $3.6$ & $54.5$ & $67.6$ \\ \cline{2-9}
      & GPT & - & - & $\mathbf{58.5}$ & $48.4$ & $10.1$ & - & - \\
      & + Data Augmentation & $88.1$ & $\mathbf{86.8}$ & $58.3$ & $\mathbf{57}$ & $\mathbf{1.3}$ & $\mathbf{63.8}$ & $\mathbf{75.7}$ \\ \hline\hline
\multirow{4}{*}{En–Pt} & M2M & - & - & $49.2$ & $36.9$ & $12.3$ & - & - \\
      & + Data Augmentation & $\mathbf{94.1}$ & $\mathbf{94.2}$ & $49.2$ & $48.3$ & $\mathbf{0.9}$ & $59.1$ & $60.1$ \\ \cline{2-9}
      & GPT & - & - & $54.1$ & $40.6$ & $13.5$ & - & - \\
      & + Data Augmentation & $92.8$ & $79.8$ & $\mathbf{54.2}$ & $\mathbf{51.5}$ & $2.7$ & $\mathbf{63.3}$ & $\mathbf{66.6}$ \\ \hline\hline
\multirow{4}{*}{En–Ru} & M2M & - & - & $29.2$ & $22.7$ & $6.5$ & - & - \\
      & + Data Augmentation & $\mathbf{86.9}$ & $\mathbf{81.1}$ & $29.3$ & $27.9$ & $\mathbf{1.4}$ & $44.6$ & $42.3$ \\ \cline{2-9}
      & GPT & - & - & $\mathbf{31.8}$ & $24.1$ & $7.7$ & - & - \\
      & + Data Augmentation & $80.1$ & $55.3$ & $31.3$ & $\mathbf{28.2}$ & $3.1$ & $\mathbf{48.8}$ & $\mathbf{49.4}$ \\ \hline\hline
\multirow{2}{*}{En–It} & M2M & - & - & $46.8$ & $34.1$ & $12.7$ & - & - \\
      & + Data Augmentation & $95.9$ & $84.6$ & $\mathbf{47}$ & $\mathbf{43.3}$ & $\mathbf{3.7}$ & $54.7$ & $50.9$ \\ \hline      
\end{tabular}
\caption{Results of the data augmentation pipeline applied to vanilla translations produced by pre-trained M2M and GPT models.}
\label{tab:non-oracle-results}
\end{table*}
}
\section{Comparison against GATE}
\label{app:gate}
For the comparison against GATE in \autoref{tab:gate-comparison}, we use exactly the same setup and metrics (Precision/Recall/F0.5) from \citet{10.1145/3600211.3604675}.
We evaluate our data augmentation pipeline on the gender re-writing task. Let's consider the \texttt{M} $\rightarrow$ \texttt{F} re-writing case: Given a source sentence with ambiguous entities, the task is to re-write an all-masculine reference translation into an all-feminine reference translation. A system might not output a re-write (in case it fails to detect any ambiguous entities or if the re-written output is the same as the input) or it might actually do a re-write. If the system performs a re-write, it’s classified as \emph{correct} if the re-write matches the all-feminine reference translation exactly. If there is any difference between the two, then the re-write is classified as \emph{incorrect}. Given these definitions, the Precision and Recall is defined as:
\begin{gather*}
\text{Precision} = \frac{\text{number of correct re-writes}}{\text{number of attempted re-writes}}\\
\text{Recall} = \frac{\text{number of correct re-writes}}{\text{total number of examples}}
\end{gather*}

\section{End-to-end MT model to generate alternatives}
\label{app:end-to-end}
We extract the bi-text used for training end-to-end models using \texttt{mtdata} \cite{gowda-etal-2021-many}. We use \texttt{sentencepiece} \cite{kudo-2018-subword} to learn a vocabulary of size $36000$ tokens. We remove sentence pairs with lengths $\geq 400$ sentencepiece tokens or exceeding a token ratio of $1\colon3$. We train all end-to-end models using the following hyper-parameters:
\begin{itemize}[noitemsep,topsep=1pt,leftmargin=.15in]
\item \texttt{batch size: 458752}
\item \texttt{decoder layers: 20}
\item \texttt{decoder layers: 3}
\item \texttt{lr: 7e-4}
\item We supervise an attention head in second from the bottom decoder layer. The scaling factor $\lambda$ for the alignment loss is set to \texttt{0.05}.
\item \texttt{embedding dim: 512}
\item shared encoder-decoder and input-output embeddings
\item \texttt{learning rate: 3e-5}
\item All reported results are gathered from a single run.
\end{itemize}
The end-end models produce gender structures without any constraints. This can result in gender structures containing phrases that differ in more than just gender inflections. To avoid this, we explicitly check the gender structures against our collected list of gender inflections and retain only those structures which pass the check.
\section{Evaluation Metrics}
\label{app:evaluation metrics}
The alternatives metrics compute the sentence level precision and recall of generating alternatives. Let $\mathbf{I}(b)$ denote an indicator function: 
\begin{equation*}
\mathbf{I}(b) = 
\begin{cases} 
      1 & b = \text{True} \\
      0 & b = \text{False}
\end{cases}
\end{equation*}
and given a sentence $x$, let $\phi(x)$ check whether $x$ contains gender structures:
\begin{equation*}
\phi(x) = 
\begin{cases} 
      \text{True} & x \text{ contains gender structures} \\ 
      \text{False} & \text{ otherwise}
\end{cases}
\end{equation*}
Let $y$ and $\hat{y}$ denote the reference from the test set and the system hypothesis respectively, 
then alternatives precision and recall can be defined as follows:
\begin{gather*}
\text{Precision} = \frac{\sum\limits_{y, \hat{y}}\mathbf{I}(\phi(y) \land \phi(\hat{y}))}{\sum\limits_{\hat{y}}\mathbf{I}(\phi(\hat{y}))} \\[2ex]
\text{Recall} = \frac{\sum\limits_{y, \hat{y}}\mathbf{I}(\phi(y) \land \phi(\hat{y}))}{\sum\limits_{y}\mathbf{I}(\phi(y))}
\end{gather*}
We compute structure metrics over the subset $S$ where both references and system outputs contain gender structures, i.e. $S=\{(y, \hat{y}) \mid \phi(y) \land \phi(\hat{y}) = \text{True}\}$. Over $S$, we compute the following statistics:
\begin{itemize}
\item Total structures: total number of gender structures present in $y$ for $(y, \hat{y}) \in S$.
\item Predicted structures: total number of gender structures present in $\hat{y}$ for $(y, \hat{y}) \in S$
\item Correct structures: total number of gender structures which are present in both $y$ and $\hat{y}$ for $(y, \hat{y}) \in S$
\end{itemize}
We can then compute structure precision and recall as follows:
\begin{gather*}
\text{Precision} = \frac{\text{Correct structures}}{\text{Predicted structures}} \\[2ex]
\text{Recall} = \frac{\text{Correct structures}}{\text{Total structures}}
\end{gather*}

\end{document}